\documentclass[journal]{IEEEtran}

\usepackage{amsmath,amssymb}
\usepackage{times}
\usepackage{lscape}
\usepackage{subfigure}
\usepackage[dvipdfmx]{graphicx}
\usepackage{algorithm,algorithmic}
\usepackage{subfigure}
\usepackage{color}
\usepackage{url}
\usepackage{xspace}

\newcommand*{\eg}{e.g.\@\xspace}
\newcommand*{\ie}{i.e.\@\xspace}

\newcommand{\bm}{\boldsymbol}
\newcommand{\ten}[1]{ \boldsymbol{\mathcal #1}}
\newcommand{\bbR}[1]{\mathbb{R}^{#1}}

\begin{document}

\title{Simultaneous Tensor Completion and Denoising by Noise Inequality Constrained Convex Optimization}

\author{Tatsuya Yokota, ~\IEEEmembership{Member,~IEEE,}
        and Hidekata Hontani, ~\IEEEmembership{Member,~IEEE}
\thanks{T. Yokota is with Nagoya Institute of Technology, 466-8555 Nagoya, Japan. e-mail: t.yokota@nitech.ac.jp (see https://sites.google.com/site/yokotatsuya/home).}
\thanks{H. Hontani is with Nagoya Institute of Technology, 466-8555 Nagoya, Japan.}
}

\markboth{Yokota \MakeLowercase{\textit{et al.}}: Simultaneous Tensor Completion and Denoising by Noise Inequality Constrained Convex Optimization}
{Yokota \MakeLowercase{\textit{et al.}}: Simultaneous Tensor Completion and Denoising by Noise Inequality Constrained Convex Optimization}

\IEEEcompsoctitleabstractindextext{
\begin{abstract}
Tensor completion is a technique of filling missing elements of the incomplete data tensors.
It being actively studied based on the convex optimization scheme such as nuclear-norm minimization.
When given data tensors include some noises, the nuclear-norm minimization problem is usually converted to the nuclear-norm `regularization' problem which simultaneously minimize penalty and error terms with some trade-off parameter.
However, the good value of trade-off is not easily determined because of the difference of two units and the data dependence.
In the sense of trade-off tuning, the noisy tensor completion problem with the `noise inequality constraint' is better choice than the `regularization' because the good noise threshold can be easily bounded with noise standard deviation.
In this study, we tackle to solve the convex tensor completion problems with two types of noise inequality constraints: Gaussian and Laplace distributions.
The contributions of this study are follows:
(1) New tensor completion and denoising models using tensor total variation and nuclear-norm are proposed which can be characterized as a generalization/extension of many past matrix and tensor completion models, 
(2) proximal mappings for noise inequalities are derived which are analytically computable with low computational complexity,
(3) convex optimization algorithm is proposed based on primal-dual splitting framework,
(4) new step-size adaptation method is proposed to accelerate the optimization, and
(5) extensive experiments demonstrated the advantages of the proposed method for visual data retrieval such as for color images, movies, and 3D-volumetric data.
\end{abstract}
}

\maketitle

\IEEEdisplaynotcompsoctitleabstractindextext

\IEEEpeerreviewmaketitle

\section{Introduction}
Completion is a technique of filling missing elements of incomplete data using the values of reference (available) elements and the structural assumptions (priors) of data.
We consider a general exact matrix/tensor completion problem as follows:
\begin{align}
  \mathop{\text{minimize}}_{\ten{X}} f(\ten{X}), \text{ s.t. } P_\Omega(\ten{X}) = P_\Omega(\ten{T}), \label{eq:general_tensor_completion_problem}
\end{align}
where $\ten{T}$ and $\ten{X} \in \bbR{I_1 \times I_2 \times \cdots \times I_N}$ are the input and output $N$-th order tensors, respectively, a cost function, $f(\cdot): \bbR{I_1 \times I_2 \times \cdots \times I_N} \rightarrow \bbR{}$, is used to evaluate (prior) structural assumptions, $P_\Omega(\ten{Z}) := \ten{Q} \circledast \ten{Z}$ with $\ten{Q} \in \{0,1\}^{I_1 \times I_2 \times \cdots \times I_N}$ is an index tensor that represents the missing and available elements of $\ten{T}$ as $0$ and $1$, respectively.
A support set, $\Omega$, is defined as $\Omega := \{(i_1, i_2, ..., i_N) \ | \  q_{i_1,i_2,...,i_N} = 1 \}$.
When the missing and available elements are independent, completion is impossible.
However, most real-world data have a few redundant properties that can be used for completion, such as symmetry, repetition, and sparsity.
When the cost function $f$ is convex and proximable, Problem \eqref{eq:general_tensor_completion_problem} can be solved by convex optimization methods such as alternating direction method of multipliers (ADMM) \cite{boyd2011distributed} and primal-dual splitting (hybrid gradient) (PDS/PDHG) method \cite{condat2013primal}.
In this paper, we refer to PDS/PDHG as PDS for simple.
For example, Problem \eqref{eq:general_tensor_completion_problem} with matrix/tensor nuclear-norm \cite{candes2009exact,candes2010power,cai2010singular,liu2009tensor,liu2013tensor}, total variation (TV) \cite{guichard1998total,dai2009physics}, and both cost functions \cite{han2014linear} have been studied.

\begin{figure}[t]
\centering
\subfigure[Lagrange form]{
\includegraphics[width=0.23\textwidth]{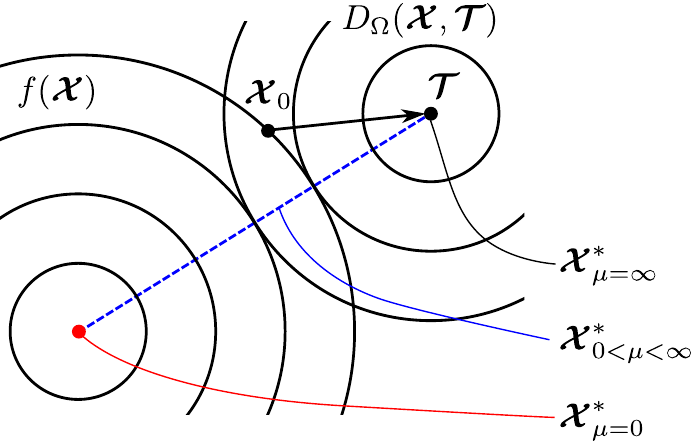}}
\subfigure[Inequality form]{
\includegraphics[width=0.23\textwidth]{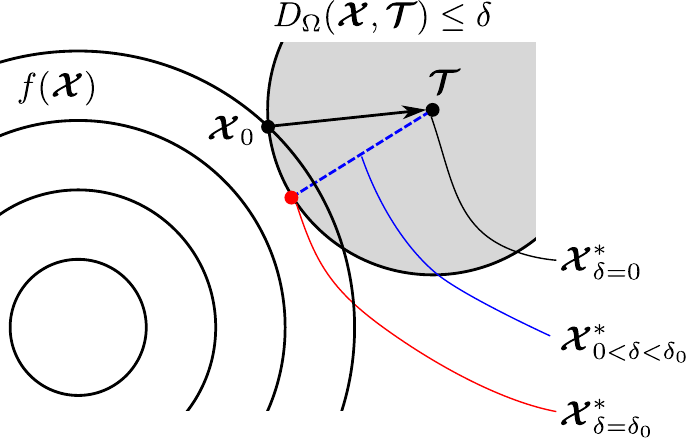}}
\caption{Geometric interpretations of convex tensor recovery and its solution.}\label{fig:geo}
\end{figure}

Next, we consider an `inexact' matrix/tensor completion problem as follows:
\begin{align}
  \mathop{\text{minimize}}_{\ten{X}} f(\ten{X}) + \mu D_\Omega(\ten{X},\ten{T}), \label{eq:inexact_tensor_completion_problem}
\end{align}
where $D_\Omega(\ten{X},\ten{T})$ is a distance measure between $\ten{X}$ and $\ten{T}$, and $\mu$ is a trade-off parameter between a prior and a distance term.
We refer to this as `Lagrange form'.
When we assume Gaussian distribution for a noise model, Problem~\eqref{eq:inexact_tensor_completion_problem} with $D_\Omega(\ten{X},\ten{T}) = || P_\Omega(\ten{X}) - P_\Omega(\ten{T}) ||_F^2$ can be considered for completion and denoising.
In similar way, Problem~\eqref{eq:inexact_tensor_completion_problem} with $D_\Omega(\ten{X},\ten{T}) = || P_\Omega(\ten{X}) - P_\Omega(\ten{T}) ||_1$
assumes Laplace distribution for a noise model.
When both functions $f$ and $D_\Omega$ are convex and proximable, it can be solved by convex optimization.
For example, Problem~\eqref{eq:inexact_tensor_completion_problem} with matrix/tensor nuclear-norm \cite{ma2011fixed,gandy2011tensor,ji2010robust}, TV \cite{rudin1992nonlinear,selesnick2012total,guo2015generalized}, and both cost functions \cite{shi2013low} have been studied.

Here, we consider `inequality form' of Problem~\eqref{eq:inexact_tensor_completion_problem} as follow:
\begin{align}
  \mathop{\text{minimize}}_{\ten{X}} f(\ten{X}), \text{s.t. } D_\Omega(\ten{X},\ten{T}) \leq \delta, \label{eq:tensor_completion_problem_inequality}
\end{align}
where $\delta$ is a noise threshold parameter.
Figure~\ref{fig:geo} illustrates the Problems~\eqref{eq:inexact_tensor_completion_problem} and \eqref{eq:tensor_completion_problem_inequality} and these solutions for different values of $\mu$ and $\delta$.
In the Lagrange form, the minimal point of convex penalty $f$ is the solution for $\mu=0$, $\ten{T}$ is the solution for $\mu=\infty$, and the solutions for $0 < \mu < \infty$ draw a curve connecting two solution points for $\mu=0$ and $\mu=\infty$.
Generally, the good trade-off exists on the curve, and it may be a projected point onto the curve from the unknown original tensor $\ten{X}_0$.
However, optimal value of $\mu$ is still unknown and has to be tuned from wide range $[0, \infty)$.
On the other hand, the range of $\delta$ can be more narrow in the inequality form when we assume the noise standard deviation $\sigma$ is known.
Let us put $\delta_0 = \sigma^2 |\Omega|$ for Gaussian and $\delta_0 = \sigma |\Omega|$ for Laplace, then optimal value of $\delta$ may exist in $[0,\delta_0]$.
If directions between the additional noise tensor and the gradient of $f$ at $\ten{X}_0$ are similar, then optimal $\delta$ would near to $\delta_0$ and $\ten{X}^*$ would also near to $\ten{X}_0$.
By contrast, if directions between the additional noise tensor and the gradient of $f$ at $\ten{X}_0$ are very different, optimal $\delta$ would be small and $\ten{X}^*$ would be far from $\ten{X}_0$.

\begin{figure}[t]
  \centering
  \includegraphics[width=0.45\textwidth]{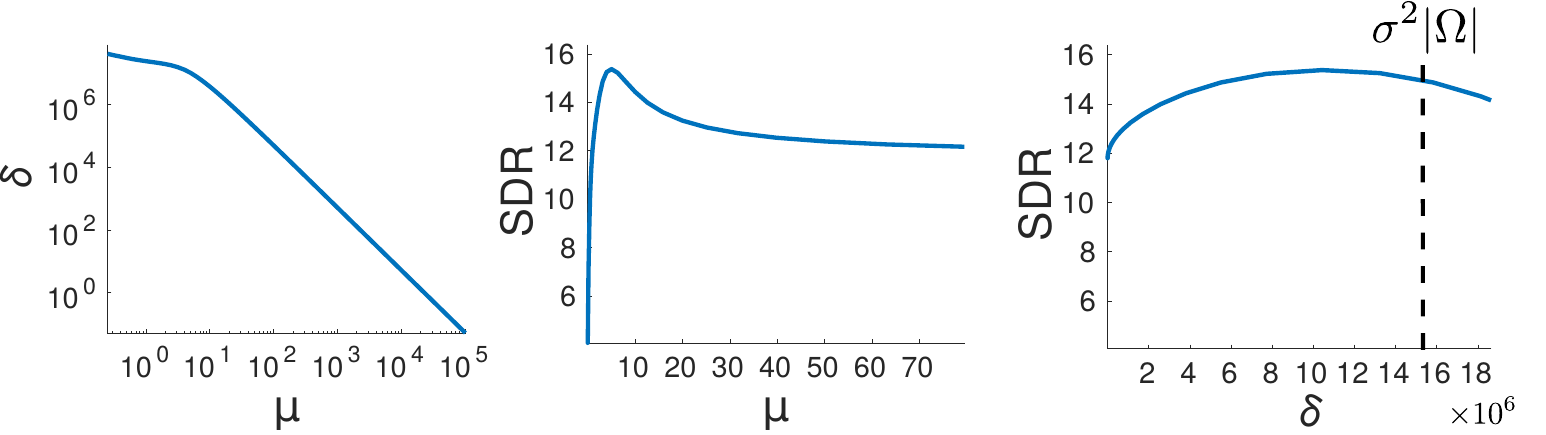}
  \caption{Examples of relationships of $\mu$ vs $\delta$ (left), $\mu$ vs signal-to-distortion ratio (SDR) (center), and $\delta$ vs SDR (right).}\label{fig:delta_mu}
\end{figure}

Problems~\eqref{eq:inexact_tensor_completion_problem} and \eqref{eq:tensor_completion_problem_inequality} are convertible with corresponding values of $\mu$ and $\delta$, however, these corresponding values of $\mu$ and $\delta$ are difficult to know.
Figure~\ref{fig:delta_mu} shows the relationship between $\delta$ and $\mu$ in a tensor completion problem with nuclear-norm and Frobenius-norm minimization.
Both Problems are linked by $\delta$ and $\mu$ which have one-to-one correspondence.

The problem here is that the inequality form is not easy to solve directly unlike Lagrange form.
In \cite{candes2010matrix}, inequality form with $\delta$ is solved by iterative optimization of Lagrange form with $\{\mu_1, \mu_2, ..., \hat{\mu} \}$ to tune optimal trade-off which corresponds to $\delta$.
This is a critical issue to solve its optimization problem with inequality using convex optimization only once.

In this study, we propose a new optimization algorithm based on PDS \cite{condat2013primal} for convex optimization problems consisting of proximable functions and noise inequality constraints.
For this purpose, we derive that the proximal mappings of noise inequality constraints based on Gaussian and Laplace distributions, which are not trivial, can be obtained using analytical calculation.
Furthermore, to accelerate the optimization, we propose a new step-size adaptation method for PDS algorithm.
For application, we define the cost function as a composition of tensor nuclear-norm and generalized TV, and conduct extensive experiments to show the advantages of the proposed methods.

Note that this work is an extension of prior study presented in conferences \cite{yokota2017simultaneous,yokota2017efficient}.
The new contributions in this study are as follows: A generalized formulation and detailed explanation of the proposed models, derivation of a new proximal mapping for Laplace noise inequality, applying a new step-size adaptation method for acceleration, and additional experiments.

The remainder of this paper is organized as follows:
In Section~\ref{sec:review}, prior studies on matrix and tensor completion methods are reviewed.
In Sections~\ref{sec:proposed} and \ref{sec:opt}, we propose a new model for tensor completion based on low rank and TV, and its optimization algorithm using a PDS approach.
In Section~\ref{sec:experiment}, we demonstrate the advantages of the proposed method over selected state-of-the-art methods using color images, movies, and 3D-volumetric images.
Lastly, we state the conclusions in Section~\ref{sec:conclusion}.

\subsection{Notations}
The notations used in this paper follow several rules.
A vector, a matrix, and a tensor are denoted by a bold lowercase letter, $\bm a \in \bbR{I}$, a bold uppercase letter, $\bm B \in \bbR{I \times J}$, and a bold calligraphic letter, $\ten{C} \in \bbR{J_1 \times J_2 \times \cdots \times J_N}$, respectively.
An $N$th-order tensor, $\ten{X} \in \bbR{I_1 \times I_2 \times \cdots \times I_N}$, can be transformed into a vector and $N$ matrix forms, which are denoted using the same character, $\bm x \in \bbR{\prod_{n=1}^N I_n}$ and $\bm X_{(n)} \in \bbR{I_n \times \prod_{k \neq n}I_k}$ for $n \in \{1, 2, ..., N\}$, respectively.
An $(i_1, i_2, ..., i_N)$-element of $\ten{X}$ is denoted by $x_{i_1,i_2,...,i_N}$ or $[\ten{X}]_{i_1,i_2,...,i_N}$.
Operator $\circledast$ represents the Hadamard product, defined as $ [\ten{X} \circledast \ten{Z}]_{i_1,i_2,...,i_N} = x_{i_1,i_2,...,i_N} z_{i_1,i_2,...,i_N} $.

\section{Review of prior works in matrix and tensor completion}\label{sec:review}

\subsection{Matrix completion models}
First, we consider a matrix completion problem as follow:
\begin{align}
  \mathop{\text{minimize}}_{\bm X} f(\bm X), \text{ s.t. } P_\Omega(\bm X) = P_\Omega(\bm T), \label{eq:matrix_completion}
\end{align}
which is a case of $N=2$ in \eqref{eq:general_tensor_completion_problem}.
For low-rank matrix completion, ideally, we want to set $f(\bm Z)=\text{rank}(\bm Z)$, however, it is NP-hard \cite{gillis2011low}.
Thus, its convex relaxation, \ie nuclear-norm, is used \cite{recht2010guaranteed}: 
\begin{align}
f(\bm Z)=|| \bm Z ||_* := \sum_{i=1}^{\min(I,J)} \sigma_i(\bm Z),
\end{align}
where $\sigma_i(\bm Z)$ is the $i$-th largest singular value of $\bm Z \in \bbR{I \times J}$.

In \cite{candes2010matrix}, a noisy case has been discussed as
\begin{align}
  \mathop{\text{minimize}}_{\bm X} ||\bm X||_*, \text{ s.t. } || P_\Omega(\bm X) -P_\Omega(\bm T) ||_F^2 \leq \delta, \label{eq:noisy_matrix_completion}
\end{align}
To solve Problem~\eqref{eq:noisy_matrix_completion}, an algorithm has been proposed, which requires solving 
\begin{align}
  \bm X_\mu^* = \mathop{\text{argmin}}_{\bm X} ||\bm X||_* + \frac{\mu}{2}|| P_\Omega(\bm X - \bm T) ||_F^2, \label{eq:nuclear_norm_mu}
\end{align}
multiple times to determine an appropriate value of $\mu > 0$ such that $|| P_\Omega(\bm X^*_{\mu} - \bm T) ||_F^2 = \delta$.
As iterative calculations of singular value decomposition are required to solve Problem~\eqref{eq:nuclear_norm_mu} \cite{ma2011fixed}, the algorithm is computationally expensive.
We refer to this algorithm as low-rank matrix completion with noise (LRMCn).

There are several studies about applications of image deblurring, denoising, and interpolation \cite{rudin1992nonlinear,vogel1998fast,guichard1998total}, where a cost function is given by TV.
The standard TV for matrix $\bm Z \in \bbR{I \times J}$ is defined by
\begin{align}
 &|| \bm Z ||_{\text{TV}} := \sum_{i,j} || \bm \nabla z_{i,j} ||_2, \label{eq:matrix_tv} \\
  & \bm \nabla z_{i,j} := \begin{pmatrix} \nabla_1 z_{i,j} \\ \nabla_2 z_{i,j}\end{pmatrix} = \begin{pmatrix} z_{i+1,j} - z_{i,j} \\ z_{i,j+1} - z_{i,j} \end{pmatrix}. 
\end{align}

Problem~\eqref{eq:matrix_completion} with the nuclear norm and TV is discussed in \cite{shi2013low}, in which it was proposed to minimize the nuclear norm using singular value thresholding and TV using gradient descent, alternately.
However, using standard gradient-based optimization is not appropriate because the nuclear norm and TV are not differentiable functions.
An alternative efficient optimization approach referred to as `proximal splitting' is gaining attention \cite{condat2013primal,bertsekas2015convex}.

\subsection{Tensor completion models}
When $N \geq 3$ in \eqref{eq:general_tensor_completion_problem}, it is not a simple extension of matrix completion because of special properties of tensors.
For example, there are two types of ranks in tensors, i.e., the canonical polyadic (CP) rank and Tucker rank \cite{kolda2009tensor}.
As the CP rank has several difficult properties, low Tucker-rank based completion is relatively well studied.

In \cite{liu2009tensor,liu2013tensor}, a case of exact tensor completion, which is Problem~\eqref{eq:general_tensor_completion_problem}, where the cost function is given by a tensor nuclear norm has been discussed, in which the tensor nuclear norm, $f_{\text{LR}}(\ten{X})$, is defined by
\begin{align}
  f_{\text{LR}}(\ten{X}) := \sum_{n=1}^N \lambda_n || \bm X_{(n)} ||_*,
\end{align}
where $\lambda_n \geq 0\ (\forall n)$ represents the weight parameters for individual tensor modes, and $\bm X_{(n)} \in \bbR{I_n \times \prod_{k \neq n} I_k}$ is the $n$-th mode unfolded matrix of tensor $\ten{X}$.
ADMM \cite{boyd2011distributed} has been employed for its minimization problem.
Furthermore, its noisy scenario has been discussed in \cite{gandy2011tensor}, which is formulated as Problem~\eqref{eq:inexact_tensor_completion_problem} with the tensor nuclear norm.
We refer to this method as low n-rank tensor completion (LNRTC).

In \cite{guo2015generalized}, a case of Problem~\eqref{eq:inexact_tensor_completion_problem}, in which the cost function is given by generalized TV (GTV) has been discussed, where GTV is defined by the sum of the generalized matrix TV of individual mode-unfolded matrices of a tensor as
\begin{align}
  f_{\text{GTV}}(\ten{X}) := \sum_{n=1}^N w_n || \bm X_{(n)} ||_{\text{GTV}},
\end{align}
where $w_n \geq 0\ (\forall n)$ represents the weight parameters for individual tensor modes, and $|| \bm Z ||_{\text{GTV}}$ for matrix $\bm Z \in \bbR{I \times J}$ is a GTV-norm, which is defined by
\begin{align}
 || \bm Z ||_{\text{GTV}} := \sum_{i,j}  \sqrt{ \sum_{\theta \in \Theta} \tau_\theta (\tilde{\nabla}_{\theta} z_{i,j})^2 },
\end{align}
where $\tau_\theta \geq 0$ represents weight parameters, $\tilde{\nabla}_{\theta}$ is the differential operator for direction $\theta$, for example, $\tilde{\nabla}_{0} z_{i,j} = z_{(i+1),j} - z_{i,j}$, $\tilde{\nabla}_{45} z_{i,j} = z_{(i+1),(j+1)} - z_{i,j}$, $\tilde{\nabla}_{90} z_{i,j} = z_{i,(j+1)} - z_{i,j}$, and $\tilde{\nabla}_{135} z_{i,j} = z_{(i-1),(j+1)} - z_{i,j}$.
This convex optimization problem is solved using the ADMM in \cite{guo2015generalized}.
We typically consider $\Theta = \{0, 90\}$ for standard matrix TV \eqref{eq:matrix_tv}. In contrast, $\Theta = \{0, 45, 90, 135\}$ is considered for GTV.
When we consider $\Theta = \{ 0 \}$ and $\tau_0 = 1$ for all $n$ in GTV, it is given by $f_{\text{GTV}}(\ten{Z}) = \sum_{n=1}^N w_n \sum_{i_1, i_2, ..., i_N} | \nabla_n z_{i_1,i_2,...,i_N} | = \sum_{i_1, i_2, ..., i_N} \sum_{n=1}^N w_n | \nabla_n z_{i_1,i_2,...,i_N} |$.
In this case, GTV is anisotropic with respect to $N$ modes in the tensors, which leads to corruption of diagonal edges.

Note that we can consider a more simple, straightforward, and isotropic tensorial extension of matrix TV, which is defined by
\begin{align}
  & f_{\text{TV}}(\ten{Z}) := \sum_{i_1, i_2, ..., i_N} || \bm \nabla z_{i_1,i_2,...,i_N} ||_{2,\bm w}, \\
  & \bm \nabla z_{i_1,i_2,...,i_N} := \begin{pmatrix} \nabla_1 z_{i_1,i_2,...,i_N} \\ \nabla_2 z_{i_1,i_2,...,i_N} \\ \vdots \\ \nabla_N z_{i_1,i_2,...,i_N} \end{pmatrix},
\end{align}
where $|| \bm v ||_{2,\bm w} := \sqrt{\sum_{n} w_n v_n^2 }$ is an weighted l2-norm, and an $n$-th mode partial differential operator is defined by $\nabla_n z_{i_1,i_2,...,i_N} := z_{i_1, ..., (i_n + 1), ..., i_N} - z_{i_1, ..., i_n, ..., i_N} $.
Instead of $f_{\mathrm{GTV}}$, we consider $f_{\mathrm{TV}}$ in this paper.

\section{Proposed model}\label{sec:proposed}
In this section, we propose a new model for tensor completion and denoising using a tensor nuclear norm and TV simultaneously.
The proposed optimization problem is given by
\begin{align}
  \mathop{\text{minimize}}_{\ten{X}} \ & \alpha f_{\text{TV}}(\ten{X}) + \beta f_{\text{LR}}(\ten{X}), \notag \\
  \text{s.t. } & v_{\min} \leq \ten{X} \leq v_{\max}, \label{eq:tensor_completion_problem}\\
 & D_\Omega(\ten{X}, \ten{T}) \leq \delta, \notag
\end{align}
where $0 \leq \alpha \leq 1$ and $\beta := 1 - \alpha$ are the weight parameters between the TV and nuclear norm terms, and the first constraint in \eqref{eq:tensor_completion_problem} imposes all values of the output tensor to be included in a range, $[v_{\min}, v_{\max}]$.
The first and second constraints are convex and these indicator functions are given by
\begin{align}
  &i_{v}(\ten{X}) := \left\{ \begin{array}{ll} 0 &  v_{\min} \leq \ten{X} \leq v_{\max} \\ \infty & \text{otherwise} \end{array} \right . , \\
  &i_{\delta}(\ten{X}) := \left\{ \begin{array}{ll} 0 & D_\Omega(\ten{X}, \ten{T}) \leq \delta \\ \infty & \text{otherwise} \end{array} \right . .
\end{align}
Using $i_{v}(\ten{X})$ and $i_{\delta}(\ten{X})$, tensor completion problem \eqref{eq:tensor_completion_problem} can be rewritten as
\begin{align}
 \mathop{\text{minimize}}_{\ten{X}}  \alpha f_{\text{TV}}(\ten{X}) + \beta f_{\text{LR}}(\ten{X}) + i_{v}(\ten{X}) + i_{\delta}(\ten{X}). \label{eq:proposed_model}
\end{align}
As these four functions are not differentiable, traditional gradient-based optimization algorithms, \eg, the Newton method, cannot be applied.
In Section~\ref{sec:opt}, we introduce and apply an efficient approach, referred to as PDS, to solve proposed optimization problem \eqref{eq:tensor_completion_problem}.

\subsection{Characterization of the proposed model}
In this section, we explain the relationship between the proposed model and prior works introduced in Section~\ref{sec:review}.
There are three characterizations of the proposed model.

First, when $N = 2$, $\alpha=0$, $\beta=1$, $\bm\lambda = [1, 0]^T$, $v_{\min} = - \infty$, and $v_{\max} = \infty$, the proposed model can be characterized as LRMCn \cite{candes2010matrix}.
In contrast with LRMCn, which solves several convex optimization problems to tune $\mu$, the proposed method can obtain its solution by solving only one convex optimization problem, and provides its tensorial extension.
Moreover, proposed model includes Laplace distribution as noise model unlike LRMCn.

Second, when $\alpha=0$, $\beta=1$, $v_{\min} = - \infty$, and $v_{\max} = \infty$, the proposed method can be characterized as LNRTC \cite{gandy2011tensor}.
In contrast with LNRTC, which employs the ADMM for solving a type of Problem \eqref{eq:inexact_tensor_completion_problem}, the proposed method employs the PDS algorithm for solving a type of Problem \eqref{eq:tensor_completion_problem_inequality}.

Third, when $\alpha=1$, $\beta=0$, $v_{\min} = - \infty$, and $v_{\max} = \infty$, the proposed model can be characterized as an isotropic version of GTV \cite{guo2015generalized}.
In contrast with GTV, in which a problem is solved using the ADMM, which requires matrix inversion through the fast Fourier transform (FFT) and inverse FFT, the proposed method does not need to consider matrix inversion.
Furthermore, the proposed method tunes the value of $\delta$ instead of $\mu$.

Additionally, our model differs from a recent work proposed in \cite{ji2016tensor} because it applies some constrained fixed-rank matrix factorization models into individual mode-matricization of a same tensor in noiseless scenario.
The problem is non-convex and it is not designed for a noise reduction model.

\section{Optimization}\label{sec:opt}
Currently, two optimization methods named ADMM and PDS have attracted attentions in signal/image processing \cite{boyd2011distributed,esser2010general,condat2013primal}.
Two optimization methods can be creatively used, for example, ADMM can be efficiently used for low-rank matrix/tensor completion \cite{lin2010augmented,chen2012matrix}, in contrast, PDS can be efficiently used for TV regularization \cite{zhu2008efficient,chambolle2011first}.
Main difference between those optimizations is that TV regularization may include a large matrix inversion in ADMM, by contrast, it can be avoided in PDS.
Thus, PDS has been used for many TV regularization methods such as TV denoising/deconvolution \cite{zhu2008efficient}, vectorial TV regularization \cite{ono2014decorrelated}, and total generalized variation in diffusion tensor imaging \cite{valkonen2013total}.

\subsection{Primal-dual splitting algorithm}\label{sec:pds}
In this section, we introduce basics of PDS algorithm.
The PDS \cite{condat2013primal} algorithm is a framework used to split an optimization problem including non-differentiable functions into several sub-optimization processes using proximal operators.
First, we consider the following convex optimization problem
\begin{align}
  \mathop{\text{minimize}}_{\bm x} f(\bm x) + h(\bm L \bm x), \label{p_problem}
\end{align}
where $f: \bbR{n} \rightarrow \bbR{}$ and $h: \bbR{m} \rightarrow \bbR{}$ are general convex functions, and $\bm L \in \bbR{m \times n}$ is a linear operator (matrix).
From the definition of convex conjugate:
\begin{align}
  h(\bm L \bm x) = \max_{\bm y} \langle \bm y, \bm L \bm x \rangle - h^*(\bm y),
\end{align}
the following saddle-point problem can be derived
\begin{align}
  \min_{\bm x} \max_{\bm y} f(\bm x) + \langle \bm y, \bm L \bm x \rangle - h^*(\bm y), \label{pd_problem}
\end{align}
where $h^*$ is a convex conjugate of $h$.
In PDS method, we focus to solve \eqref{pd_problem} instead of \eqref{p_problem}.

For optimality of $(\widehat{\bm x}, \widehat{\bm y})$, at least the following conditions are satisfied:
\begin{align}
   &\bm 0 = \bm p(\widehat{\bm x},\widehat{\bm y}) \in \partial f(\widehat{\bm x}) + \bm L^T \widehat{\bm y}, \label{opt_cond_x} \\
   &\bm 0 = \bm d(\widehat{\bm x},\widehat{\bm y}) \in \partial h^*(\widehat{\bm y}) - \bm L \widehat{\bm x}, \label{opt_cond_y}
\end{align}
where $\partial$ stands for sub-gradient of functions, and we consider primal and dual residual vectors as some $\bm p$ and $\bm d$, respectively.

Based on sub-gradient descent/ascent method, natural update rules are given by
\begin{align}
  \bm x^{k+1} = \bm x^k - \gamma_1 \Delta \bm x^k, \\
  \bm y^{k+1} = \bm y^k + \gamma_2 \Delta \bm y^k,
\end{align}
where $\Delta \bm x^k \in \bm L^T \bm y^k + \partial f(\bm x^k)$ and $\Delta \bm y^k \in \bm L \bm x^{k+1} - \partial h^*(\bm y^k)$ are update directions, and $\gamma_1$ and $\gamma_2$ are step size parameters.

When $f$ and $h$ are proximable functions, proximal gradient method can be introduced as
\begin{align}
  &\bm x^{k+1} = \text{prox}_{\gamma_1 f} [ \bm x^k - \gamma_1 \bm L^T \bm y^k], \label{update_x} \\
  &\bm y^{k+1} = \text{prox}_{\gamma_2 h^*} [ \bm y^k + \gamma_2 \bm L \bm x^{k+1}], \label{update_y}
\end{align}
where proximal mapping is defined by
\begin{align}
  \text{prox}_{\lambda g}[\bm z] := \mathop{\text{argmin}}_{\bm u} \lambda g(\bm u) + \frac{1}{2}|| \bm u - \bm z ||_2^2.
\end{align}
Note that let us put $\bm z^* =  \text{prox}_{\lambda g}[\bm z]$, then we have 
\begin{align}
  \bm z - \bm z^* \in \lambda \partial g(\bm z^*). \label{property_of_prox}
\end{align}
Optimization algorithm using \eqref{update_x} and \eqref{update_y} is called as ``Arrow-Hurwicz method'' which is the original version of PDS method.
A generalization of PDS method is given by
\begin{align}
  &\bm x^{k+1} = \text{prox}_{\gamma_1 f} [ \bm x^k - \gamma_1 \bm L^T \bm y^k], \label{update_x_pds} \\
  &\bm y^{k+1} = \text{prox}_{\gamma_2 h^*} [ \bm y^k + \gamma_2 \bm L (\bm x^{k+1} + \theta (\bm x^{k+1} - \bm x^{k}) )], \label{update_y_pds}
\end{align}
where $\theta \in [0, 1]$ is a some scalar.
Usually, $\theta = 1$ is chosen because that the fast convergence $\mathcal{O}(1/N)$ of $\theta = 1$ is theoretically proved in contrast to  $\mathcal{O}(1/\sqrt{N})$ of $\theta = 0$ \cite{chambolle2011first}.
Therefore, Formulations \eqref{update_x_pds}-\eqref{update_y_pds} are recognized as a standard version of PDS method, currently.
Nonetheless, Arrow-Hurwicz method is practically competitive with standard PDS method that a experimental report exists in \cite{chambolle2011first}.

Using Moreau decomposition rule:
\begin{align}
  \bm z = \text{prox}_{\lambda g*}[\bm z] + \lambda \text{prox}_{\frac{1}{\lambda} g}\left[\frac{1}{\lambda} \bm z\right],
\end{align}
Update rule \eqref{update_y_pds} can be rewritten by 
\begin{align}
  &\widetilde{\bm y}^{k+1} =  \bm y^k + \gamma_2 \bm L (\bm x^{k+1} + \theta (\bm x^{k+1} - \bm x^{k}) ), \label{y_update_2_1} \\
  &\bm y^{k+1} = \widetilde{\bm y}^{k+1} - \gamma_2 \text{prox}_{ \frac{1}{\gamma_2}h} \left[\frac{1}{\gamma_2} \widetilde{\bm y}^{k+1}\right]. \label{y_update_2_2}
\end{align}
If proximal mapping of $h^*$ is more difficult to calculate or derive than that of $h$, then update rule \eqref{y_update_2_1}-\eqref{y_update_2_2} is a convenient choice to implement.

Thus, PDS can be applied into many convex optimization problems that the proximal mappings of $f$ and $h$ are given as analytically computable operations.
Furthermore, above formulation can be easily extended into the composite optimization of multiple convex functions $h_j(\bm L_j \bm x)$ \cite{condat2013primal}.


\subsection{Proposed algorithm}
In this section, we apply the PDS algorithm to the proposed optimization problem.
Introducing dual variables $\ten{U} \in \bbR{I_1 \times I_2 \times \cdots \times I_N}$, $\bm Y = [\bm y_1, \bm y_2, ..., \bm y_N] \in \bbR{\prod_{n=1}^N I_n \times N}$, and $\{\ten{Z}^{(n)} \in \bbR{I_1 \times I_2 \times \cdots \times I_N} \}_{n=1}^N$, Problem \eqref{eq:proposed_model} can be rewritten as
\begin{align}
   \mathop{\text{minimize}}_{\bm x} \ \ & i_\delta(\ten{X}) + i_v(\ten{U}) \notag \\
                                      & + \alpha || \bm Y^T ||_{2,1} + \beta \sum_{n=1}^N \lambda_n || \bm Z_{(n)}^{(n)} ||_*, \label{eq:problem_lrtv_pds}\\
   \text{s.t. } \ \ & \ten{U} = \ten{X}, \ \bm y_n = \sqrt{ w_n } \bm D_n \bm x \ (\forall n), \notag \\ 
                    & \ten{Z}^{(n)} = \ten{X} \ (\forall n), \notag
\end{align}
where $\bm x \in \bbR{\prod_{n=1}^N I_n}$ is the vectorized form of $\ten{X}$ and $\bm D_n$ is the linear differential operator of the $n$-th mode of the tensor.
$|| \cdot ||_{2,1}$ is the $l_{2,1}$-norm of the matrix, defined as $|| \bm Z ||_{2,1} := \sum_{j=1}^J || \bm z_j ||_2 $ for matrix $\bm Z = [\bm z_1, \bm z_2, ..., \bm z_J]\in \bbR{I \times J}$.
Algorithm~\ref{alg:lrtv_pds} can be derived using the PDS framework in Problem \eqref{eq:problem_lrtv_pds}.
We refer to this algorithm as the ``low-rank and TV (LRTV)--PDS'' algorithm.

Note that the $l_{2,1}$-norm, the nuclear norm, and $i_v$ are clearly proximable functions whose calculations are given by
\begin{align}
  &\text{prox}_{\gamma ||\cdot||_{2,1}}(\bm Z) = \left[ \text{prox}_{\gamma ||\cdot||_2}(\bm z_1), ..., \text{prox}_{\gamma ||\cdot||_2}(\bm z_J) \right], \label{eq:prox_l21} \\
  &\text{prox}_{\gamma ||\cdot||_2}(\bm z) = \frac{\bm z}{||\bm z||_2} \max(||\bm z||_2-\gamma,0), \\
  &\text{prox}_{\gamma ||\cdot||_*}(\bm Z) = \bm U \max(\bm\Sigma - \gamma,0) \bm V^T, \label{eq:prox_nuc} \\
  &\text{prox}_{i_v} (\ten{Z}) = \max(\min(\ten{Z},v_{\max}),v_{\min}), \label{eq:prox_D}
\end{align}
where $\bm s = [s_1, ..., s_I]^T \in \bbR{I}$ with $s_i = \max\left( 1 - \frac{\gamma}{|| \bm z_i ||_2}, 0 \right)$ for $\bm Z = [\bm z_1, \bm z_2, ..., \bm z_I]^T \in \bbR{I \times J}$, and $(\bm U, \bm \Sigma, \bm V)$ are the left, center-diagonal, and right matrices, respectively, of the singular value decomposition of $\bm Z$.
Proximal mappings of $i_\delta$ with Gaussian and Laplace noise models are provided by Sections~\ref{sec:proof_l2} and~\ref{sec:proof_l1}.

\begin{algorithm}[t]
\caption{LRTV--PDS algorithm}\label{alg:lrtv_pds}
\begin{algorithmic}[1]
  \STATE {\bf input} : $\ten{T}$, $\ten{Q}$, $\delta$, $v_{\min}$, $v_{\max}$, $\alpha$, $\bm w$, $\beta$, $\bm \lambda$, $\gamma_1$, $\gamma_2$;
  \STATE {\bf initialize} : $\ten{X}^0$, $\ten{U}^0$, $\bm Y^0$, $\ten{Z}^{(n)0}$ $(\forall n)$, $k = 0$;
  \REPEAT
    \STATE $\bm v \leftarrow \bm u^k + \sum_{n=1}^N \bm z^{(n)k} + \sum_{n=1}^N \sqrt{ w_n } \bm D_n^T \bm y_n^k$ ;
    \STATE $\bm x^{k+1} = \text{prox}_{i_\delta} \left[ \bm x^k - \gamma_1 \bm v \right] $;
    \STATE $\bm h \leftarrow 2 \bm x^{k+1} - \bm x^k$;
    \STATE $\tilde{\bm u} \leftarrow \bm u^k + \gamma_2 \bm h$;
    \STATE $\bm u^{k+1} = \tilde{\bm u} - \gamma_2 \text{prox}_{i_v} \left[ \frac{1}{\gamma_2} \tilde{\bm u} \right]$;
    \STATE $\widetilde{\bm Y} \leftarrow \bm Y^k + \gamma_2 [\sqrt{ w_1 } \bm D_1 \bm h, ..., \sqrt{ w_N } \bm D_N \bm h]$;
    \STATE $\bm Y^{k+1} = \widetilde{\bm Y} - \gamma_2 \text{prox}_{\frac{\alpha}{\gamma_2}||\cdot||_{2,1}} \left[ \frac{1}{\gamma_2} \widetilde{\bm Y} \right]$;
    \STATE $\widetilde{\bm Z}^{(n)} \leftarrow \bm Z_{(n)}^{(n)k} + \gamma_2 \bm H_{(n)}$; $(\forall n)$
    \STATE $\bm Z_{(n)}^{(n)k+1} = \widetilde{\bm Z}^{(n)} - \gamma_2 \text{prox}_{\frac{\beta \lambda_n}{\gamma_2}||\cdot||_*} \left[ \frac{1}{\gamma_2}\widetilde{\bm Z}^{(n)} \right]$; $(\forall n)$
    \STATE $k \leftarrow k + 1$;
  \UNTIL convergence
\end{algorithmic}
\end{algorithm}

\subsection{Proximal mapping of Gaussian noise inequality} \label{sec:proof_l2}
In this section, we consider the following problem:
\begin{align}
  \mathop{\text{min}}_{\ten{X}} \frac{1}{2}|| \ten{Z} - \ten{X} ||_F^2 \text{ s.t. } || \ten{Q} \circledast (\ten{T} - \ten{X}) ||_F^2 \leq \delta.
\end{align}
Focusing on the elements $q_{i_1,i_2,...,i_N} = 0$, those elements are independent with respect to inequality.
Thus, for minimizing each element cost $( z_{i_1,i_2,...,i_N} - x_{i_1,i_2,...,i_N} )^2$, we obtain
\begin{align}
  x^*_{i_1,i_2,...,i_N} = z_{i_1,i_2,...,i_N} \ \ \ (q_{i_1,i_2,...,i_N} = 0). \label{eq:for_q_zero}
\end{align}

\begin{figure}[t]
\centering
\includegraphics[width= 0.35\textwidth]{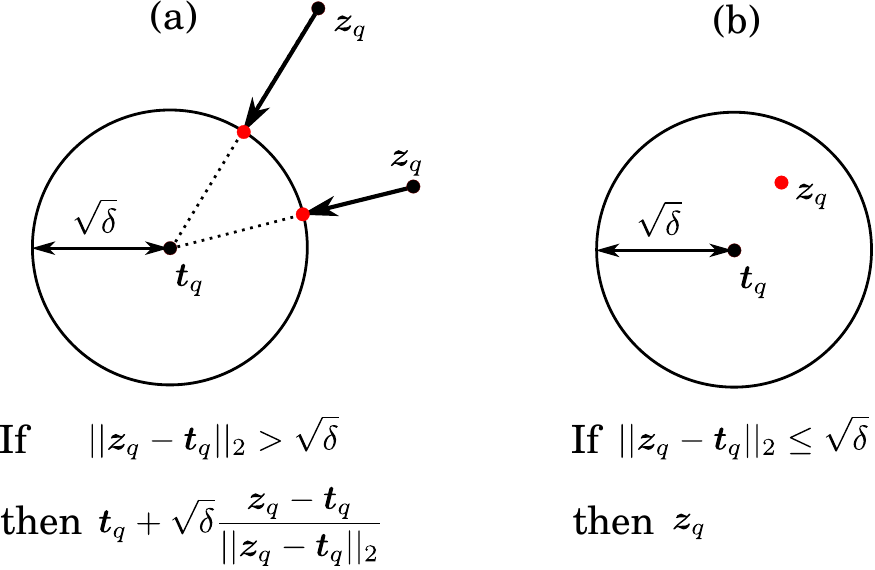}
\caption{Projection onto sphere.  It can be calculated analytically.}\label{fig:projection_to_sphere}
\end{figure}

Focusing on the elements $q_{i_1,i_2,...,i_N} = 1$, the optimization problem is given as 
\begin{align}
 \mathop{\text{minimize}}_{\bm x_q}  \frac{1}{2}|| \bm z_q - \bm x_q ||_2^2 \text{ s.t. } ||\bm t_q - \bm x_q||_2^2 \leq \delta, \label{eq:prox_ind_qone}
\end{align}
where $\bm z_q$, $\bm t_q$, and $\bm x_q$ are vectors consisting of all elements of $\ten{Z}$, $\ten{T}$, and $\ten{X}$, respectively, that satisfy $q_{i_1,i_2,...,i_N} = 1$.
The solution of \eqref{eq:prox_ind_qone} is given by a projection of $\bm z_q$ on the sphere with center $\bm t_q$ and radius $\sqrt{\delta}$, or by the $\bm z_q$ that is in that sphere (see Fig.~\ref{fig:projection_to_sphere}).
We can consider two cases: (a) $|| \bm z_q - \bm t_q ||_2  > \sqrt{\delta}$, and (b) $|| \bm z_q - \bm t_q ||_2  \leq \sqrt{\delta}$.
Thus, we obtain
\begin{align}
  \bm x_q^* &= \bm t_q + \min\left( 1, \eta \right) (\bm z_q - \bm t_q) \notag\\
           &=  [ 1 - \min( 1, \eta ) ] \bm t_q + \min( 1, \eta ) \bm z_q \notag \\
           &= \max(0,1-\eta) \bm t_q + [1 - \max(0,1-\eta)] \bm z_q, \label{eq:for_q_one}
\end{align}
where $\eta = \frac{\sqrt{\delta}}{|| \bm z_q - \bm t_q ||_2} =  \frac{\sqrt{\delta}}{|| \ten{Q} \circledast(\ten{Z} - \ten{X}) ||_2}$.
Combining \eqref{eq:for_q_zero} and \eqref{eq:for_q_one}, we obtain
\begin{align}
  x_{i_1, i_2, ..., i_N}^* =& q_{i_1, i_2, ..., i_N} \max(0,1-\eta) t_{i_1, i_2, ..., i_N} \notag \\
                     & + [1 - q_{i_1, i_2, ..., i_N}\max(0,1-\eta)] z_{i_1, i_2, ..., i_N},
\end{align}
and the proximal mapping of $i_\delta$ with Gaussian noise model is given by
\begin{align}
  \text{prox}_{i_\delta} (\ten{Z}) = \widetilde{\ten{Q}}\circledast \ten{T} + (1-\widetilde{\ten{Q}})\circledast \ten{Z}, \label{eq:prox_delta}
\end{align}
where $\widetilde{\ten{Q}} = \max (0,1-\eta) \ten{Q}$.
Clearly, this computational complexity is linear with respect to the size of tensor $\ten{X}$.

\begin{figure}[t]
\centering
\includegraphics[width= 0.35\textwidth]{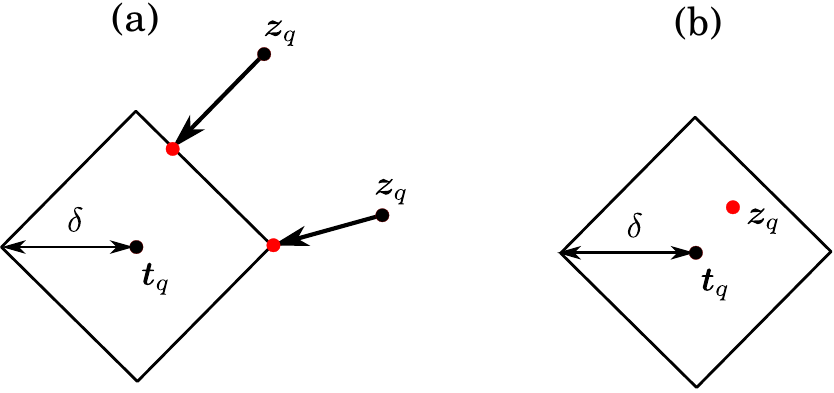}
\caption{Projection onto polyhedron.  It can not be calculated analytically.}\label{fig:polyhedron}
\end{figure}

\subsection{Proximal mapping of Laplace noise inequality}\label{sec:proof_l1}
In this section, we consider the following problem:
\begin{align}
  \mathop{\text{min}}_{\ten{X}} \frac{1}{2}|| \ten{Z} - \ten{X} ||_F^2 \text{ s.t. } || \ten{Q} \circledast (\ten{T} - \ten{X}) ||_1 \leq \delta.
\end{align}
In the same way to \eqref{eq:for_q_zero}, focusing on the elements $q_{i_1,i_2,...,i_N} = 0$, we obtain
\begin{align}
  x^*_{i_1,i_2,...,i_N} = z_{i_1,i_2,...,i_N} \ \ \ (q_{i_1,i_2,...,i_N} = 0). \label{eq:for_q_zero_l1}
\end{align}

Focusing on the elements $q_{i_1,i_2,...,i_N} = 1$, the optimization problem is given as 
\begin{align}
 \mathop{\text{minimize}}_{\bm x_q}  \frac{1}{2}|| \bm z_q - \bm x_q ||_2^2 \text{ s.t. } ||\bm t_q - \bm x_q||_1 \leq \delta, \label{eq:prox_ind_qone_l1}
\end{align}
The solution of \eqref{eq:prox_ind_qone_l1} is given by a projection of $\bm z_q$ on the polyhedron with center $\bm t_q$, or the $\bm z_q$ that is in that polyhedron (see Fig.~\ref{fig:polyhedron}).
However, it can not be calculated analytically unlike Gaussian noise model.
Generally, it is resolved by linear search problem.
Let us consider Lagrange form of \eqref{eq:prox_ind_qone_l1} as 
\begin{align}
  \mathop{\text{minimize}}_{\bm x_q}  \frac{1}{2}|| \bm z_q - \bm x_q ||_2^2 + \tau ||\bm t_q - \bm x_q||_1, \label{eq:lagrange_l1_projection}
\end{align}
then the solution of \eqref{eq:lagrange_l1_projection} can be given by soft-thresholding:
\begin{align}
  \widehat{\bm x}_\tau = \bm t_q + \bm s \circledast \max(\bm h - \tau, 0),
\end{align}
where $s_i := \text{sign}[ (z_q)_i - (t_q)_i ]$, and $h_i := | (z_q)_i - (t_q)_i |$.
The linear search problem can be given by
\begin{align}
  \tau^* = \mathop{\text{argmin}}_{\tau \geq 0} \tau, \text{ s.t. } || \widehat{\bm x}_\tau - \bm t_q ||_1 \leq \delta. \label{eq:linear_search}
\end{align}
Finally, the solution of \eqref{eq:prox_ind_qone_l1} is given as $\bm x_q^* = \widehat{\bm x}_{\tau^*}$.

\begin{figure}[t]
  \centering
  \includegraphics[width=0.45\textwidth]{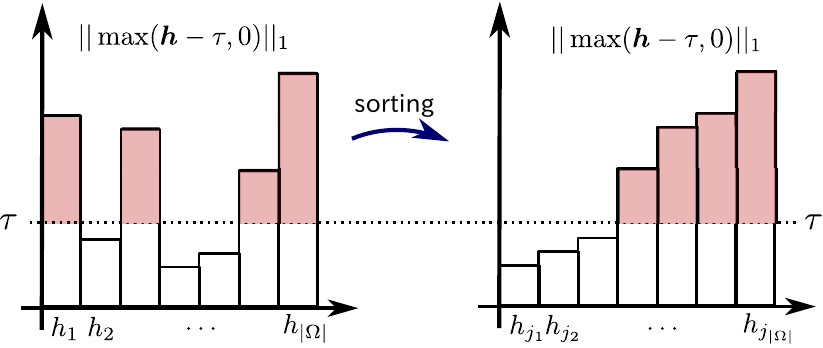}
  \caption{Illustration of $||\max(\bm h-\tau,0)||$ with sorting.}\label{fig:left_part}
  \vspace{3mm}
  \centering
  \includegraphics[width=0.49\textwidth]{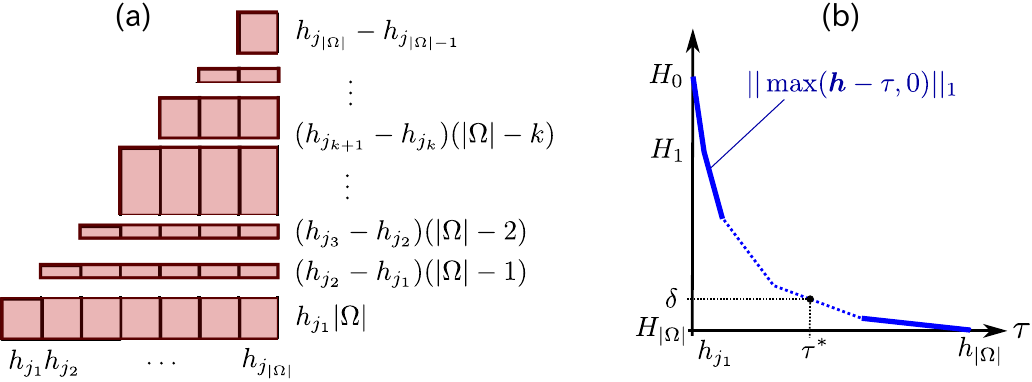}
  \caption{Reconstruction of $||\max(\bm h-\tau,0)||$: (a) block representations, (b) variation with respect to $\tau$.}\label{fig:left_part_2}
\end{figure}

Next, we show an efficient algorithm to solve \eqref{eq:linear_search}.
The computational complexity of the proposed algorithm is only $\mathcal{O}(|\Omega| \log |\Omega|)$ for sorting.
First, the constraint in \eqref{eq:linear_search} can be rewritten by
\begin{align}
  || \bm s \circledast \max(\bm h-\tau,0) ||_1 = \sum_i \max(h_i-\tau,0) \leq \delta. \label{eq:l1_constraint}
\end{align}
Next, we consider the sorting of $h_i$ as
\begin{align}
  h_{j_1} \leq h_{j_2} \leq \cdots \leq h_{j_{|\Omega|}}.
\end{align}
The left part of \eqref{eq:l1_constraint} can be illustrated by Fig.~\ref{fig:left_part}.
Obviously, the left part is $ H_0 := ||\bm h||_1$ with $\tau = 0$.
Next, when $\tau = h_{j_1}$, the left part can be obtained by
\begin{align}
   H_1 := H_0 - h_{j_1}|\Omega|.
\end{align}
In general, for $\tau = h_{j_{k+1}}$, the left part can be calculated by
\begin{align}
   H_{k+1} = H_{k} - (h_{j_{k+1}} - h_{j_k})(|\Omega| - k).
\end{align}
Fig.~\ref{fig:left_part_2} helps us to understand above formulations.
We have $H_0 \geq H_1 \geq H_2 \geq \cdots \geq H_{|\Omega|}=0$.
Thus, we can find $k'$ that satisfies $H_{k'} \geq \delta \geq H_{k'+1}$ by linear computational complexity of $\mathcal{O}(|\Omega|)$.
Finally, optimal value of $\tau$ can be given by
\begin{align}
  \tau^* = h_{j_{k'+1}} - \frac{\delta - H_{k'+1}}{|\Omega|-k'}.
\end{align}

\subsection{Step-size adaptation}
From the theory of fixed point algorithm, global convergence of PDS method with sufficiently small step-size has been proven \cite{esser2010general,chambolle2011first,he2012convergence}.
However, small step-size leads usually slow convergence, and optimal step-size may not be constant, i.e., it may adaptively change in optimization process.
Furthermore, appropriate balance between primal and dual step-size parameters is not trivial.
To tackle this issue, Goldstein {\em et al.} has been proposed a nice adaptation rule of primal-dual step-size parameters in 2015 \cite{goldstein2015adaptive}.
By inheriting and improving Goldstein's rule work, we proposed a new step-size adaptation rule.

In this section, we consider to apply above step-size adaptation rules into our new tensor completion model.
It is very important for step-size adaptation to consider primal and dual residual vectors:
\begin{align}
  \bm p^{k+1} &:= \frac{1}{\gamma_1}(\bm x^k - \bm x^{k+1}) - (\bm u^k - \bm u^{k+1}) \notag \\
               & - \sum_n \left\{ \sqrt{w_n}\bm D_n^T(\bm y_n^k - \bm y_n^{k+1}) + (\bm z^{(n)k} - \bm z^{(n)k+1}) \right\}, \\
  \bm d^{k+1} &:= \begin{pmatrix}
                  \frac{1}{\gamma_2}(\bm u^k - \bm u^{k+1}) - (\bm x^k - \bm x^{k+1}) \\
                  \frac{1}{\gamma_2}(\bm y_1^k - \bm y_1^{k+1}) - \sqrt{w_1}\bm D_1(\bm x^k - \bm x^{k+1}) \\
                   \vdots \\
                  \frac{1}{\gamma_2}(\bm y_N^k - \bm y_N^{k+1}) - \sqrt{w_N}\bm D_N(\bm x^k - \bm x^{k+1}) \\
                  \frac{1}{\gamma_2}(\bm z^{(1)k} - \bm z^{(1)k+1}) - (\bm x^k - \bm x^{k+1}) \\
                   \vdots \\
                  \frac{1}{\gamma_2}(\bm z^{(N)k} - \bm z^{(N)k+1}) - (\bm x^k - \bm x^{k+1})
                 \end{pmatrix}.
\end{align}
Primal and dual vectors $(\bm p^{k+1},\bm d^{k+1})$ can be derived based on Eqs.~\eqref{opt_cond_x}, \eqref{opt_cond_y}, and \eqref{property_of_prox}.

\begin{figure}[t]
  \centering
  \subfigure[Variations of residual]{
  \includegraphics[width=0.45\textwidth]{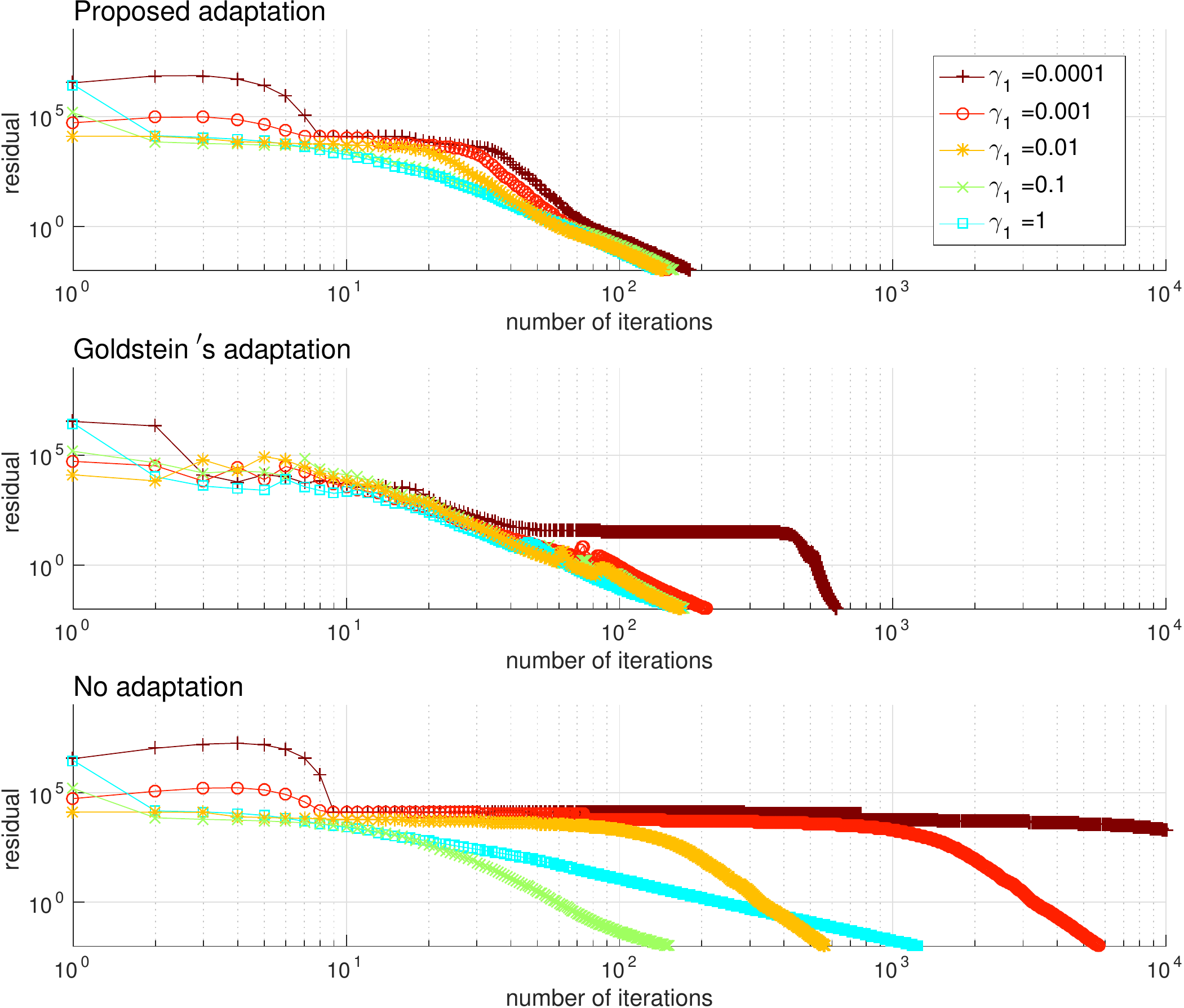}}
  \subfigure[Variations of primal-dual rate]{
  \includegraphics[width=0.45\textwidth]{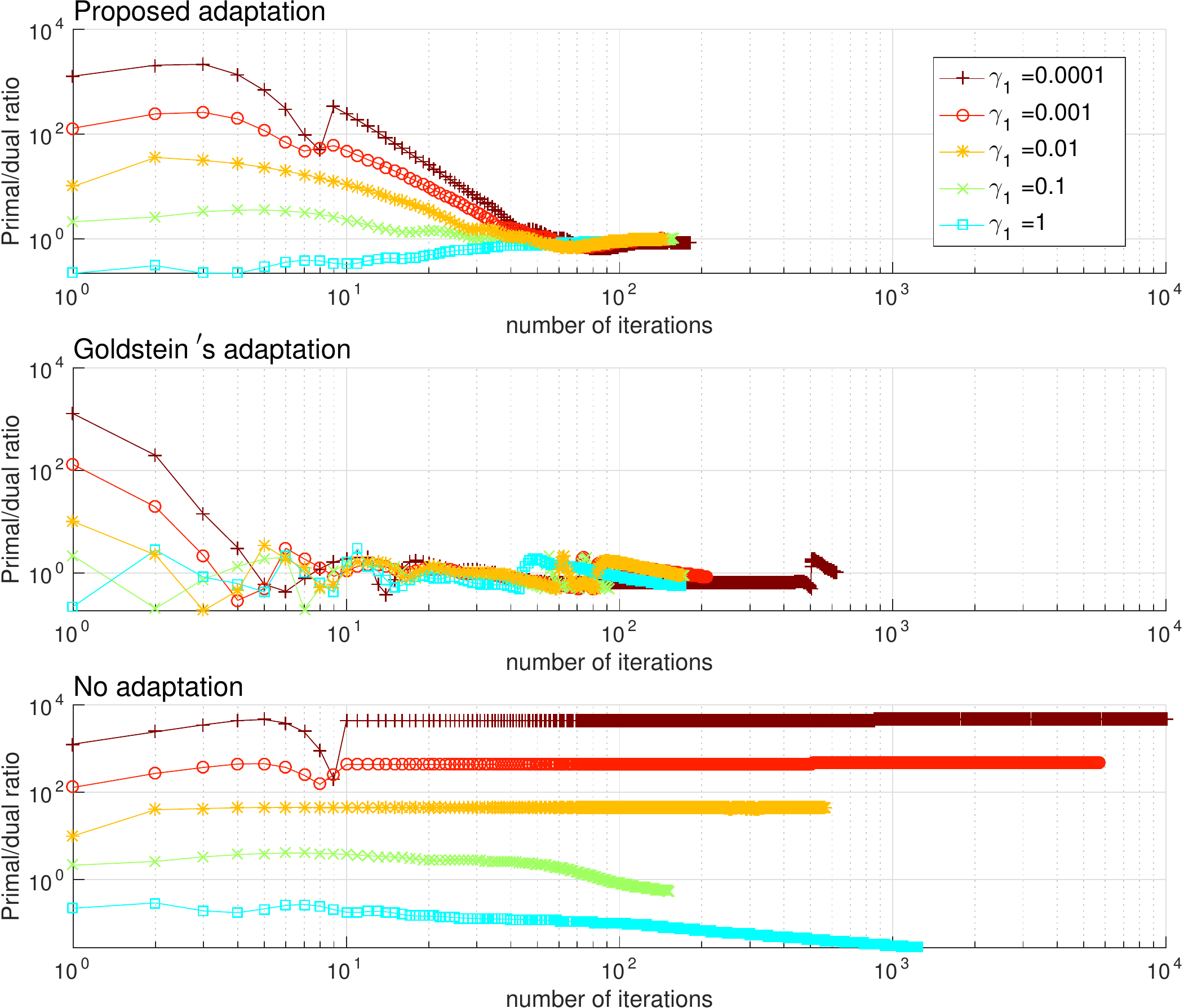}}
  \caption{Convergence behaviors of residual and primal-dual rate with different initializations of $(\gamma_1, \gamma_2)$. We set $\gamma_2=1/(8\gamma_1)$.}\label{fig:convergence}
\end{figure}

\subsubsection{Goldstein's rule \cite{goldstein2015adaptive}}
Here, we introduce an adaptation rule for primal-dual step-size of PDS proposed by Goldstein {\em et al.} in 2015.
In order to balance $||\bm p^{k+1}||_2$ and $||\bm d^{k+1}||_2$, primal and dual step-size $(\gamma_1,\gamma_2)$ are adjusted as follow:
\begin{itemize}
  \item If $||\bm p^{k+1}||_2 \geq 2 ||\bm d^{k+1}||_2$, then $\gamma_1 \leftarrow \gamma_1/(1-\alpha)$, $\gamma_2 \leftarrow \gamma_2(1-\alpha)$, and $\alpha \leftarrow \alpha \eta$,
  \item If $||\bm d^{k+1}||_2 \geq 2 ||\bm p^{k+1}||_2$, then $\gamma_2 \leftarrow \gamma_2/(1-\alpha)$, $\gamma_1 \leftarrow \gamma_1(1-\alpha)$, and $\alpha \leftarrow \alpha \eta$,
\end{itemize}
where $\alpha \in [0,1)$, and $\eta \in (0,1)$.
Moreover, to prevent too large step-size, a backtracking condition is defined as
\begin{align}
  B_{k+1} := & \frac{c}{2\gamma_1}|| \bm x^{k+1} - \bm x^k ||^2  + \frac{c}{2\gamma_2}|| \bm v^{k+1} - \bm v^k ||^2 \notag \\
           & -2 (\bm v^{k+1} - \bm v^k)^T \bm A (\bm x^{k+1}-\bm x^k),
\end{align}
where $c \in (0,1)$ is a constant (typically $c=0.9$), and 
\begin{align}
  &\bm v = \left[ \bm u^T, \bm y_1^T, ..., \bm y_N^T, \bm z^{(1)T}, ..., \bm z^{(N)T} \right]^T, \\
  &\bm A^T = \left[ \bm I, \sqrt{w_1}\bm D_1^T, ..., \sqrt{w_N}\bm D_N^T, \bm I, ..., \bm I \right],
\end{align}
and if $B_{k+1} < 0$, then $\gamma_1 \leftarrow \gamma_1/2$ and $\gamma_2 \leftarrow \gamma_2/2$.

\subsubsection{Proposed rule}
In contrast that Goldstein's rule balances the primal and dual step-sizes based on inequality condition, the proposed rule is based on primal-dual ratio $R_{k+1} := \frac{||\bm p^{k+1}||_2}{||\bm d^{k+1}||_2} $ as follow:
\begin{align}
  &\gamma_1 \leftarrow \gamma_1 R_{k+1}^{\rho_\gamma}, \\
  &\gamma_2 \leftarrow \gamma_2 R_{k+1}^{-\rho_\gamma},
\end{align}
where $\rho_{\gamma} \in (0,1)$ is a positive small scalar (typically $\rho_\gamma=0.05$).

In addition, the proposed technique has acceleration and backtracking rules based on the angle between current and next descent directions:
\begin{align}
  &w_P^{k+1} := \frac{\langle \bm x^k - \bm x^{k+1}, \bm p^{k+1} \rangle}{||\bm x^k - \bm x^{k+1}||_2 \cdot || \bm p^{k+1} ||_2}, \\
  &w_D^{k+1} := \frac{\langle \bm v^k - \bm v^{k+1}, \bm d^{k+1} \rangle}{||\bm v^k - \bm v^{k+1}||_2 \cdot || \bm d^{k+1} ||_2}.
\end{align}
The acceleration and backtracking rules are given as follow:
\begin{itemize}
  \item If $w_P^{k+1} \geq 0.9$, then $\gamma_1 \leftarrow \beta \gamma_1$;
  \item If $w_D^{k+1} \geq 0.9$, then $\gamma_2 \leftarrow \beta \gamma_2$;
  \item If $w_P^{k+1} \leq 0$, then $\gamma_1 \leftarrow \zeta \gamma_1$;
  \item If $w_D^{k+1} \leq 0$, then $\gamma_2 \leftarrow \zeta \gamma_2$;
\end{itemize}
where $\beta > 1$ and $\zeta \in (0,1)$ are positive scalars (typically $\beta=1.01$ and $\zeta=0.9$).



\begin{figure}[t]
  \centering
  \includegraphics[width=0.49\textwidth]{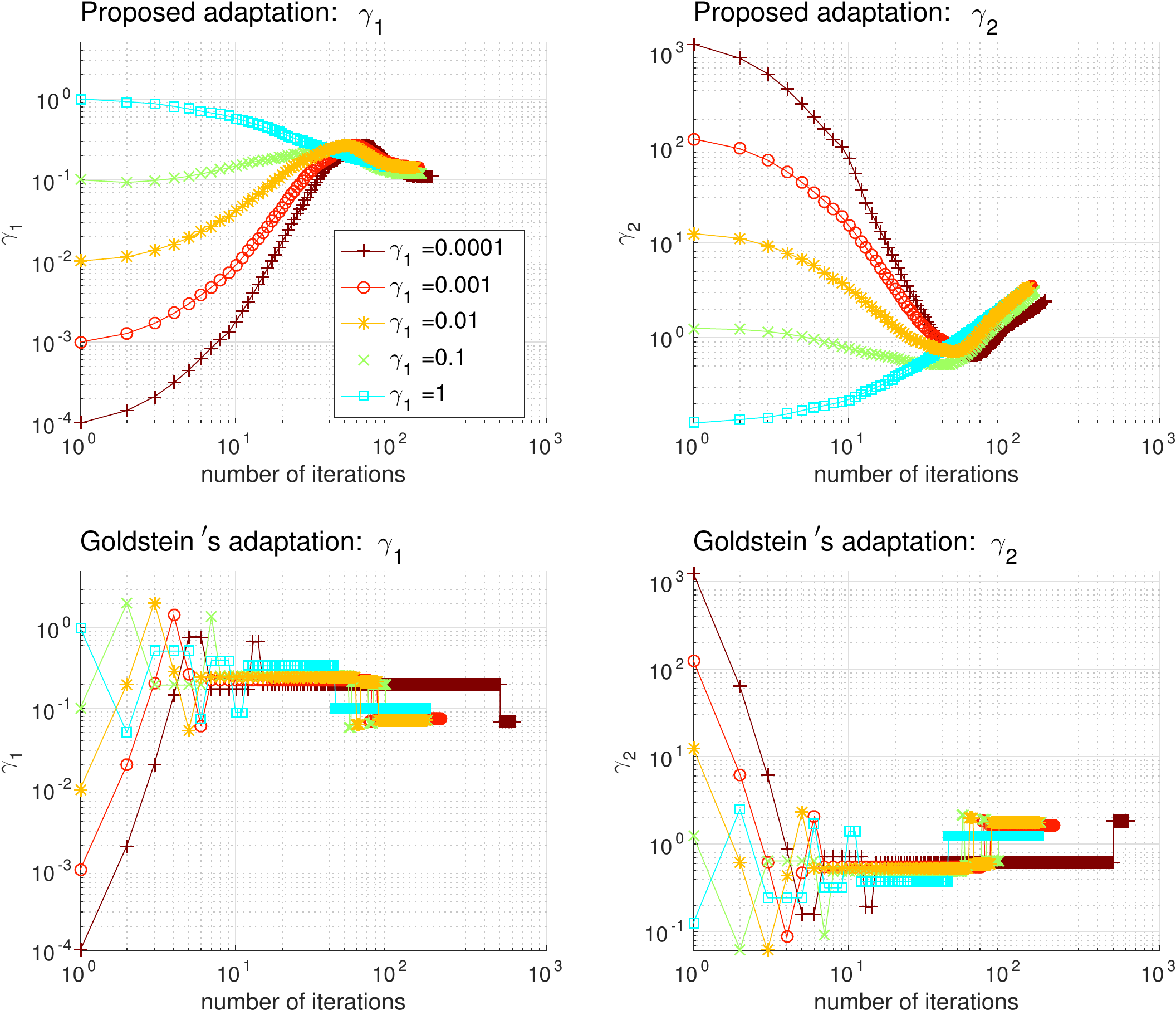}
  \caption{Variations of $(\gamma_1,\gamma_2)$ adapted by Goldstein's and proposed rules with different initializations of $(\gamma_1, \gamma_2)$.}\label{fig:convergence_gamma}
\end{figure}

\section{Experiments}\label{sec:experiment}

\begin{figure}[t]
\centering
\includegraphics[width=0.49\textwidth]{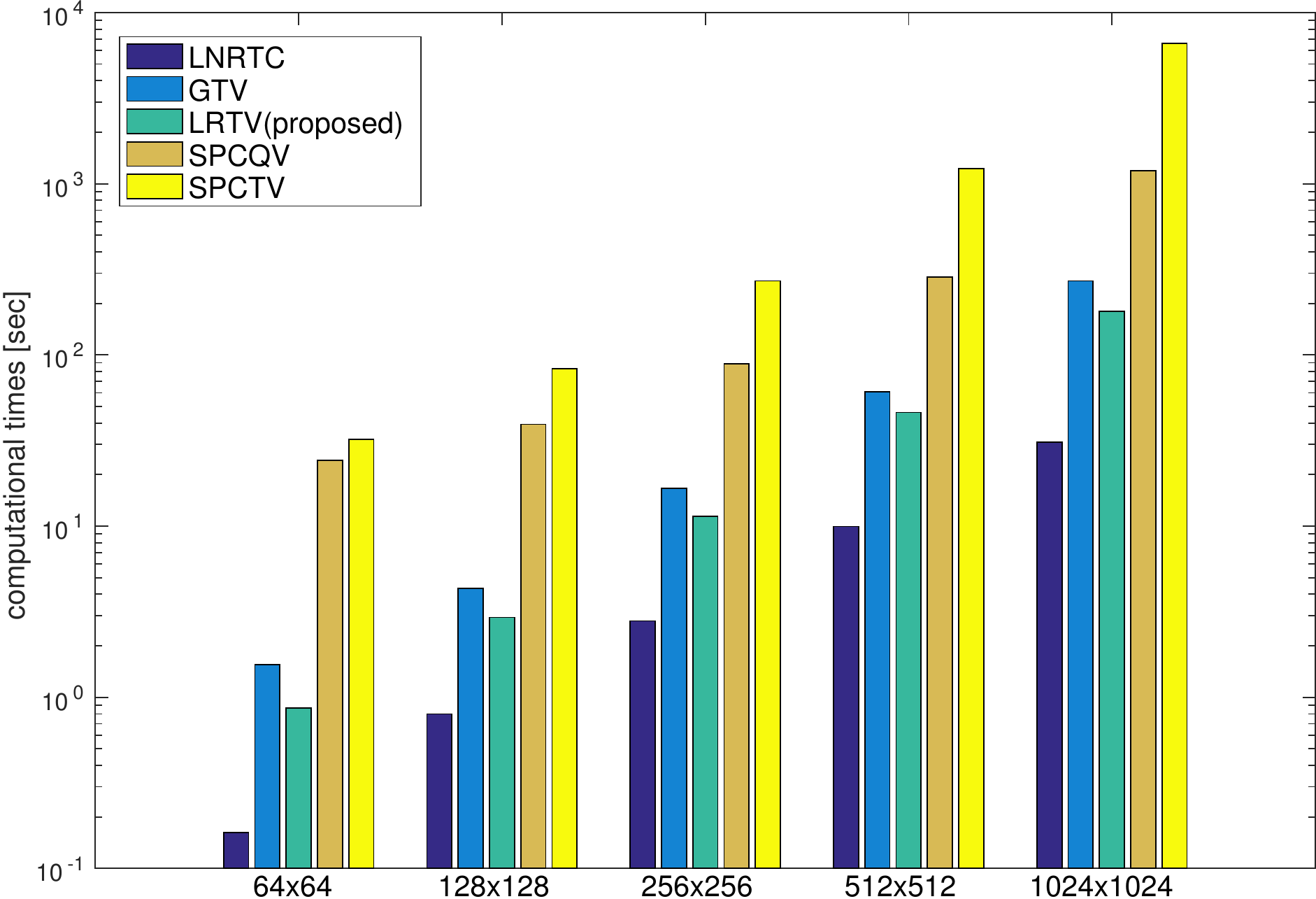}
\caption{Computational times of various methods.}\label{fig:comptimes}
\end{figure}

\subsection{Evaluation of optimization}
In this experiments, we show convergence behaviors of the proposed optimization algorithm for comparing two step-size adaptation techniques: Goldstein's adaptation rule \cite{goldstein2015adaptive}, and the proposed adaptation rule.
We applied the proposed LRTV-PDS algorithm to missing and noisy image recovery problem with the proposed step-size adaptation, Goldstein's adaptation, and no adaptation for various initial-step sizes $\gamma_1 \in \{10^{-4}, 10^{-3}, 10^{-2}, 10^{-1}, 1\}$ and $\gamma_2 = 1/(8\gamma_1)$.
For comparison, a consistent stopping criterion is employed as $ ||\bm p^{k+1}||_2^2 + ||\bm d^{k+1}||_2^2 \leq 10^{-2} $.
Figure~\ref{fig:convergence} shows the convergence behaviors of primal-dual residual and residual ratio for three adaptations with different $(\gamma_1, \gamma_2)$.
Note that graphs are illustrated by log-scale.
In the case of no adaptation, convergence speed is widely varied with different step-sizes, and primal-dual rates are not controlled.
In the case of Goldstein's adaptation, convergence speed is accelerated and primal-dual rates are controlled well.
However, convergence is sometimes still slow with some wrong initial step-sizes.
In the proposed adaptation, the convergence is more stably accelerated with different initial step-sizes and primal-dual rates are more smoothly controlled than Goldstein's adaptation.
Figure~\ref{fig:convergence_gamma} shows the convergence behaviors of $\gamma_1$ and $\gamma_2$ by two adaptation methods for various initializations.
We can see that $\gamma_1$ and $\gamma_2$ are converged almost same values for all initializations in both adaptation methods.
However, the proposed method control ($\gamma_1$, $\gamma_2$) more smoothly than Goldstein's method.

\subsection{Computational times}
In this experiments, we compare computational times of the proposed method with typical convex methods and state-of-the-art non-convex tensor completion methods.
For convex optimization, we selected LNRTC \cite{gandy2011tensor} and GTV \cite{guo2015generalized} for comparison.
In contrast, for non-convex optimization (tensor decomposition), we selected smooth parallel factor (PARAFAC) tensor completion with total/quadratic variation (SPCTV/SPCQV) \cite{yokota2016smooth}.
In contrast with LNRTC, which is based on a low Tucker-rank prior, SPCTV/SPCQV is based on a low CP-rank prior.
We used the SPCTV/SPCQV MATLAB toolbox, which is distributed by IEEEXPLORE\footnote{\url{http://ieeexplore.ieee.org/document/7502115/media}}.
As packages for LNRTC and GTV might not be distributed, we implemented these algorithms in MATLAB.
Please note that the LNRTC and GTV are optimized by ADMM.
For the ADMM, a penalty parameter $\mu_c$ should be adaptively controlled for fast convergence.
According to \cite{bertsekas2014constrained}, increasing update of $\mu_c^{k+1} \geq \mu_c^{k}$ is recommended for fast convergence.
In \cite{lin2010augmented,gandy2011tensor,guo2015generalized}, a kind of increasing update rule $\mu_c^{k+1} = \rho_c \mu_c^{k}$ is employed for ADMM based matrix/tensor completion problems, where $\rho_c > 1$.
Thus, we updated the penalty parameter of ADMM by $\mu_c^{k+1} = 1.1 \mu_c^{k}$ for LNRTC and GTV.
Figure~\ref{fig:comptimes} shows computational times [sec] of all comparison methods for various image sizes.
Note that the vertical axis of the graph is depicted by log-scale.
Convex methods were clearly faster than non-convex methods.
The proposed method was relatively faster than GTV, and slower than LNRTC.
Thus, the proposed LRTV-PDS algorithm is computationally efficient in consideration of that the proposed method is a generalization of LNRTC and GTV, and solving the noise inequality constrained problems, which is more complicated than regularization problems.

\begin{figure}[t]
\centering
\includegraphics[width=0.05\textwidth]{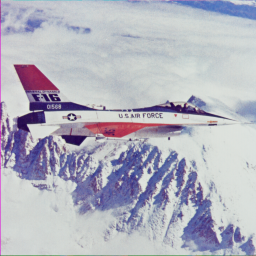}
\includegraphics[width=0.05\textwidth]{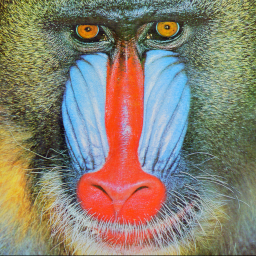}
\includegraphics[width=0.05\textwidth]{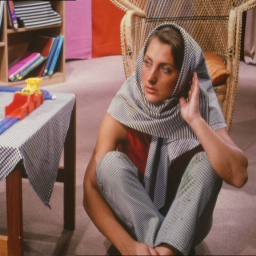}
\includegraphics[width=0.05\textwidth]{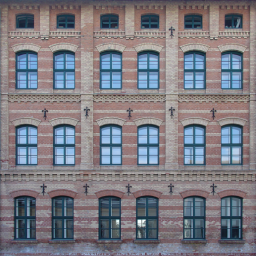}
\includegraphics[width=0.05\textwidth]{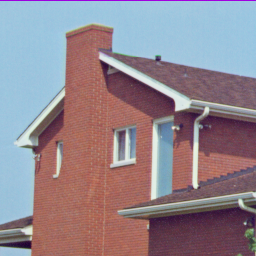}
\includegraphics[width=0.05\textwidth]{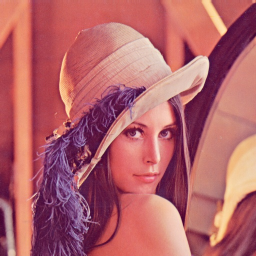}
\includegraphics[width=0.05\textwidth]{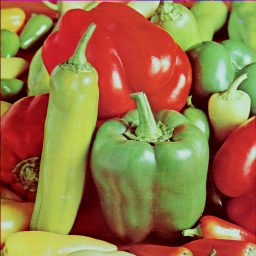}
\includegraphics[width=0.05\textwidth]{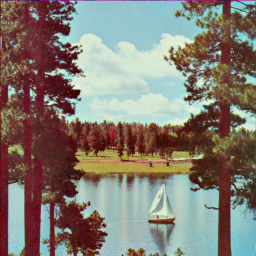}
\caption{Test color images: `airplane', `baboon', `barbara', `facade' (left to right in the first row), `house', `lena', `peppers', `sailboat' (left to right in the second row).}\label{fig:test_images}
\end{figure}

\begin{figure*}[t]
\centering
\subfigure[PSNR color map]{
\includegraphics[width=0.79\textwidth]{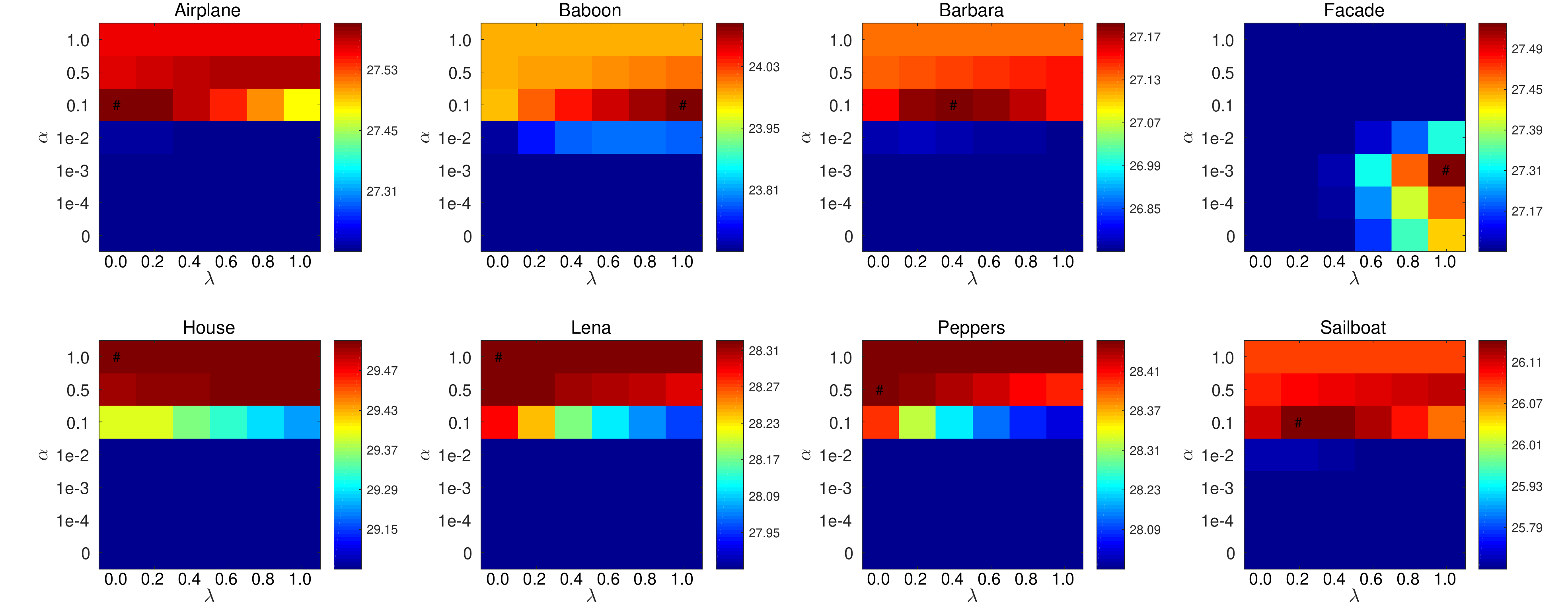}}
\subfigure[SSIM color map]{
\includegraphics[width=0.79\textwidth]{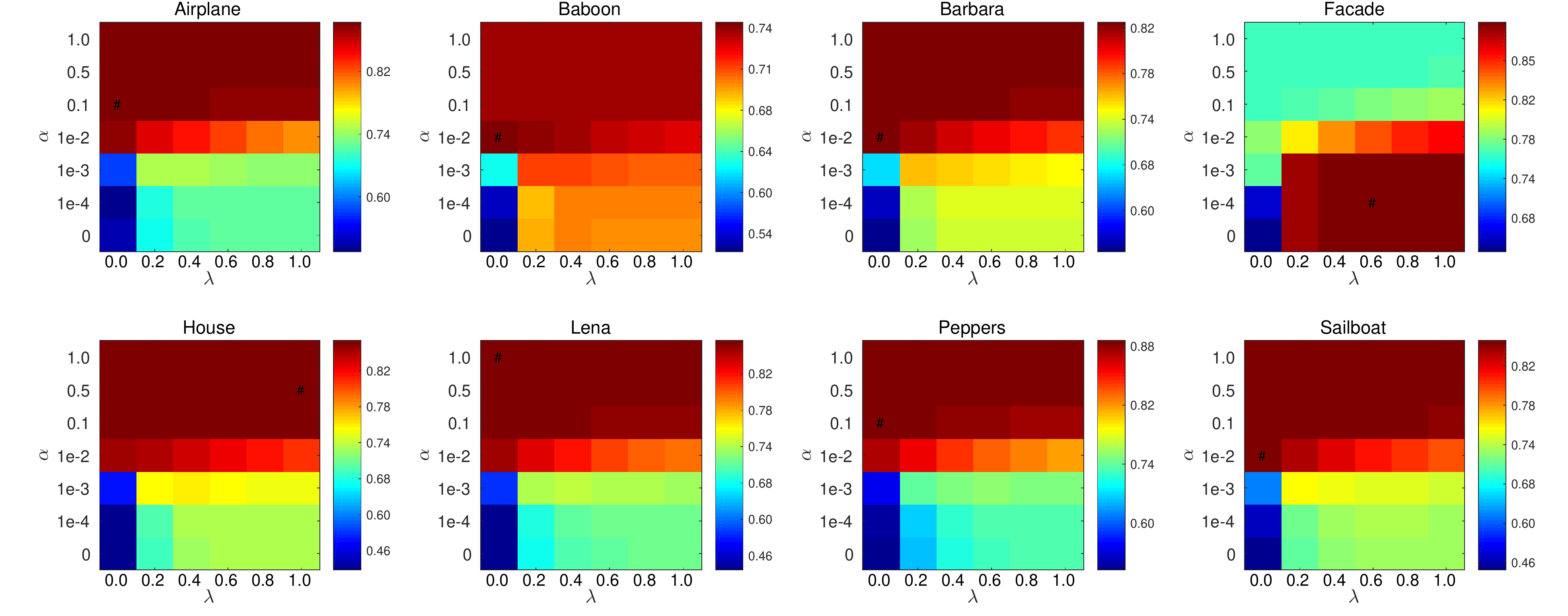}}
\caption{PSNR and SSIM color maps for various values of hyperparameters $\alpha$ and $\lambda$ using eight test color images with 30\% missing elements. `\#' is placed on the maximum values of PSNR in each colormap.}\label{fig:map}
\end{figure*}

\subsection{Color image recovery: trying various parameters}
In this experiments, we tried various trade-off parameter settings of the proposed methods.
Figure~\ref{fig:test_images} shows the test images used in this experiment.
Eight images have a size of (256$\times$256$\times$3).
All images were corrupted by removing arbitrary voxels and adding Gaussian noise, $\mathcal{N}(0,20)$.
Missing ratios were considered as $\rho = 0.3$.
For the proposed method, we tuned the values of $\alpha$, $\lambda$, and $\delta$ to evaluate the peak signal to noise ratio (PSNR) and structural similarity (SSIM).
Other parameters were considered as $\beta = 1-\alpha$, $\bm w = [0.5, 0.5, 0.0]^T$, and $\bm \lambda = [\lambda/2, \lambda/2, 1-\lambda]^T$.
Figure~\ref{fig:map} shows the color illustration of the PSNR and SSIM values for various combinations of $\alpha$ and $\lambda$ in individual images.
The most suitable combinations of $\alpha$ and $\lambda$ depend on the images.
Three images, referred to as `house', `lena', and `peppers', prefer TV regularization (large $\alpha$). In contrast, `facade' prefers nuclear norm regularization (small $\alpha$).
Other images, referred to as `airplane', `baboon', `barbara', and `sailboat', prefer intermediate mixing of TV and nuclear norm regularizations.

\subsection{Color image recovery: robustness to the noise level in comparison with state-of-the-art methods}\label{sec:noise}
In this experiments, we evaluated the robustness of the proposed method to the different levels of noises which were generated by Gaussian and Laplace distributions with $\sigma \in \{10, 20, 30, 40\}$.
In addition, 30\% voxels were removed randomly.
We generated missing and noisy image dataset by using eight benchmark images for 10 realizations.
Thus, totally $640 = 8 (\text{images}) \times 4 (\text{noise levels}) \times 2 (\text{Gaussian/Laplace}) \times 10 (\text{realizations})$ missing and noisy images were generated.
For the Laplace noise, the proposed method and GTV have options of the projection onto polyhedron and soft-thresholding, respectively.
We applied LNRTC, GTV, the proposed method, SPCQV, and SPCTV with appropriate parameter settings for all 640 corrupted images.
For LNRTC, we tuned a regularization parameter $\mu_{\mathrm{lr}} \in \{ 10^{-5}, 10^{-4}, ..., 10^{5} \}$ for each image.
For GTV, we tuned a regularization parameter $\mu_{\mathrm{tv}} \in \{ 10^{-5}, 10^{-4}, ..., 10^{5} \}$ for each image.
We applied GTV with $l_{2,1}$-TV regularization for images with Gaussian noise, and GTV with $l_1$-TV regularization for images with Laplace noise.
For the proposed method, we tuned a noise threshold $\delta = \rho_\delta \delta_0$ for $\rho_\delta \in \{0.1, 0.2, 0.3, ..., 1.0\}$.
Assuming $\sigma$ and noise distribution is known in advance, we set $\delta_0 := \sigma^2 |\Omega|$ for Gaussian noise distribution and $\delta_0 := \sigma |\Omega|$ for Laplace noise distribution.
Balancing parameter $(\alpha,\beta)$ between nuclear-norm and total variation were decided based on the results shown in Figure~\ref{fig:map}.
For SPCQV/SPCTV, we manually tuned smoothness parameters because of expensive computational costs of non-convex optimizations.

\begin{figure*}[t]
  \centering
  \subfigure[30\% missing with additive Gaussian noise]{
  \includegraphics[width=0.84\textwidth]{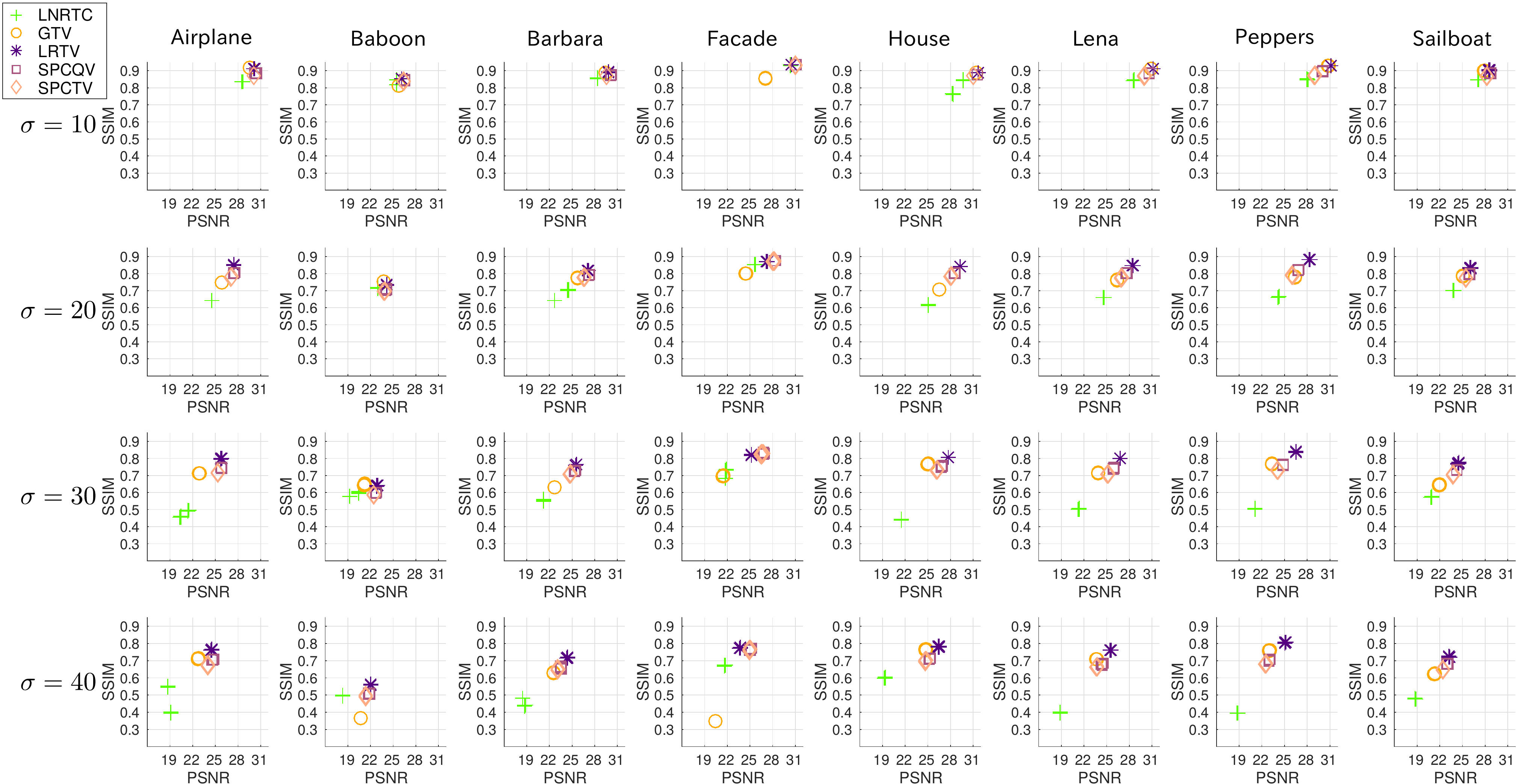}}
  \subfigure[30\% missing with additive Laplace noise]{
  \includegraphics[width=0.84\textwidth]{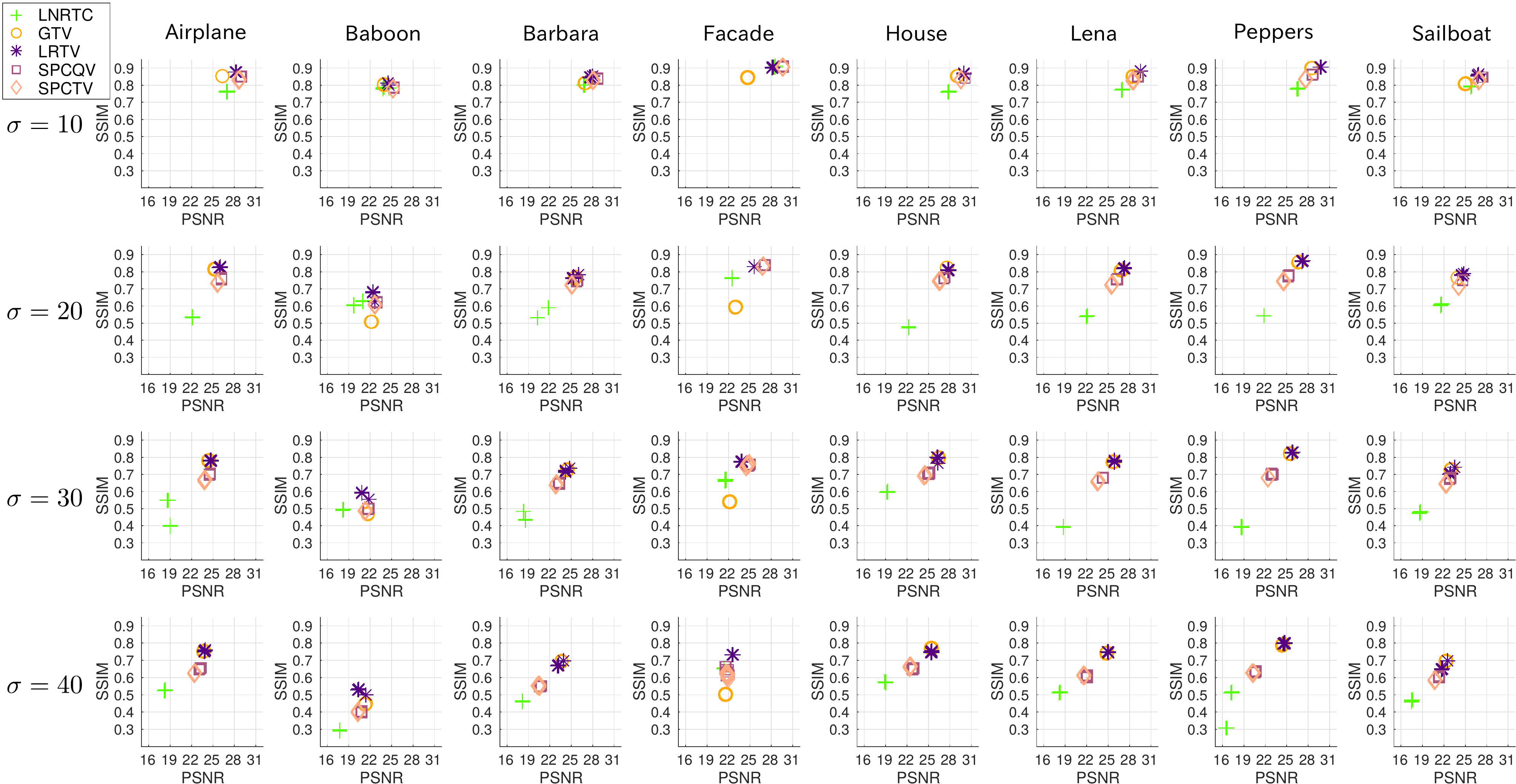}}
  \caption{PSNR and SSIM plots in color image completion and denoising.}\label{fig:PSNR_SSIM_color_image}
\end{figure*}

For each images and methods, we computed (PSNR,SSIM) as a point in two dimensional visual quality measure space.
Figure~\ref{fig:PSNR_SSIM_color_image} shows results of (PSNR,SSIM) scatter plots for individual noise settings with images.
We can see that the proposed LRTV method were relatively robust to strong noise compared with other methods in both Gaussian and Laplace noise settings excepting ``Facade''.

\subsection{Color image recovery: robustness to the missing rate in comparison with state-of-the-art methods}
Next, we evaluated the robustness of the proposed method to the different levels of missing rate $\rho \in \{10\%, 30\%, 50\%, 70\%, 90\% \}$.
Totally, $400 = 8 (\text{images}) \times 5 (\text{missing rates}) \times 10 (\text{realizations})$ images were generated for this experiments.
Gaussian noises with $\sigma = 20$ were added for all missing images.
In the same way to Section~\ref{sec:noise}, we applied LNRTC, GTV, the proposed method, SPCQV, and SPCTV for all corrupted images.

\begin{figure*}[t]
  \centering
  \includegraphics[width=0.84\textwidth]{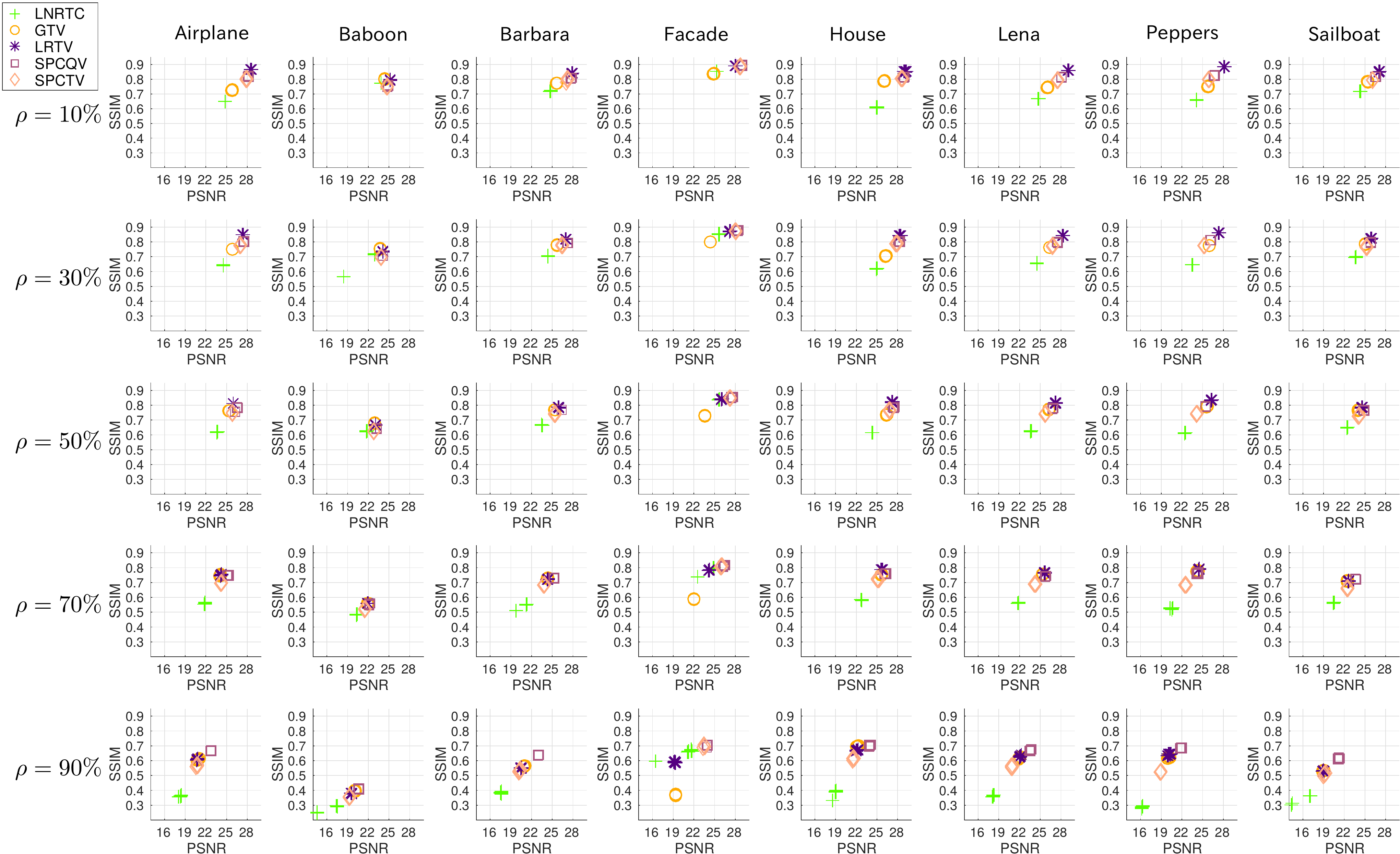}
  \caption{PSNR and SSIM plots in color image completion and denoising for various missing rates.}\label{fig:PSNR_SSIM_color_image_missing}
\end{figure*}

Figure~\ref{fig:PSNR_SSIM_color_image_missing} shows results of (PSNR,SSIM) scatter plots for individual missing rates with images.
We can see that the proposed LRTV outperformed the other methods for low-missing rates: $\rho \in \{10\%, 30\% \}$.
The proposed method and SPCQV were competitive for middle-missing rate: $\rho = 50\%$, and SPCQV outperformed the other methods for high-missing rates: $\rho \in \{70\%, 90\%\}$.

\begin{table*}[t]
\caption{PSNR/SSIM/computational time[sec] comparison in color movie (4D tensor) completion and denoising}\label{tab:psnr_movie}
\centering
\setlength{\tabcolsep}{3pt}
\begin{tabular}{l l l l l l l } \hline
  Missing rate & LNRTC          & GTV              & LRTV(proposed)  & SPCQV            &     SPCTV   \\ \hline
 10\%     & 28.35/.9003/{\bf 8}  & 30.27/.9284/1119 & {\bf 30.86}/{\bf .9335}/268 & 30.02/.9051/3004 & 30.02/.9083/3090 \\
 30\%     & 26.79/.8498/{\bf 11} & 27.54/.9068/1138 & 28.27/{\bf .9134}/372 & 29.64/.9005/3709 & {\bf 29.65}/.9042/3711 \\
 50\%     & 24.97/.8208/{\bf 9}  & 25.26/.8714/1183 & 26.10/.8818/450 & 28.98/.8919/4311 & {\bf 29.01}/{\bf .8966}/4284 \\ \hline
\end{tabular}
\end{table*}

\begin{table*}[t]
\caption{SDR/SSIM/computational time[sec] comparison for 3D-MR image completion and denoising}\label{tab:mri}
\centering
\setlength{\tabcolsep}{3pt}
\begin{tabular}{l l l l l l l} \hline
 name   &  Missing rate & LNRTC                & GTV                    & LRTV(proposed)  & SPCQV            &     SPCTV   \\ \hline
 citrus &      10\%     & 23.84/.8980/{\bf 7}  & 25.42/{\bf .9211}/218  & {\bf 25.61}/{\bf .9211}/33  & 24.06/.8921/987  & 23.45/.8897/1360 \\
 citrus &      30\%     & 20.91/.8626/{\bf 10} & 23.01/.9079/245        & {\bf 23.36}/{\bf .9082}/47  & 22.57/.8760/1224 & 21.84/.8744/1546 \\
 citrus &      50\%     & 18.07/.8364/{\bf 9}  & 20.40/.8805/237        & {\bf 20.85}/{\bf .8869}/62  & 20.83/.8544/1301 & 19.96/.8516/1609 \\
 citrus &      70\%     & 15.01/.7771/{\bf 8}  & 17.37/{\bf .8524}/242  & 17.88/.8521/74              & {\bf 18.76}/.8237/1179 & 17.68/.8181/1466 \\
 citrus &      90\%     & 10.52/.6814/{\bf 10} & 13.66/{\bf .7810}/229  & 14.06/.7808/130             & {\bf 15.66}/.7696/935  & 14.02/.7415/1278 \\ \hline
 tomato &      10\%     & 26.21/.9129/{\bf 11} & 27.93/.9270/224        & {\bf 28.10}/{\bf .9276}/37  & 26.96/.9078/957  & 27.06/.9130/1240 \\
 tomato &      30\%     & 24.49/.9064/{\bf 10} & 26.79/.9211/238        & {\bf 27.22}/{\bf .9212}/50  & 26.12/.8982/1029 & 26.02/.9036/1424 \\
 tomato &      50\%     & 22.13/.8755/{\bf 9}  & 25.18/.9133/236        & {\bf 25.97}/{\bf .9134}/61  & 25.03/.8864/1025 & 24.66/.8911/1528 \\
 tomato &      70\%     & 18.93/.8385/{\bf 8}  & 22.62/.8992/239        & {\bf 23.65}/{\bf .9002}/68  & 23.44/.8768/845  & 22.44/.8720/1441 \\
 tomato &      90\%     & 13.45/.7860/{\bf 11} & 18.02/{\bf .8716}/232  & 19.11/.8708/111             & {\bf 20.07}/.8380/672  & 17.94/.8218/1248 \\ \hline
\end{tabular}
\end{table*}

\subsection{Color movie recovery}
In this experiments, we evaluated the proposed method using color movie completion and denoising.
A data set, referred to as `bootstrap', is distributed by Microsoft Research\footnote{\url{http://research.microsoft.com/en-us/um/people/jckrumm/wallflower/testimages.htm}} \cite{toyama1999wallflower}.
We corrupted it by removing arbitrary voxels and adding Gaussian noise, $\mathcal{N}(0,10)$.
The input movie (4D tensor) has a size of 120 pixels $\times$ 160 pixels $\times$ 3 color-channels $\times$ 100 frames.
We considered $\bm w = [0.4, 0.4, 0.0, 0.2]^T$ and $\bm \lambda = [0.2, 0.2, 0.2, 0.4]^T$, and $(\alpha,\beta,\delta)$ were tuned manually.
Table~\ref{tab:psnr_movie} shows the values of PSNR, SSIM, and computational time [sec] for various missing ratios and for all comparison methods.
In computational times, LNRTC was the fastest, and the proposed method was the second.
In accuracy, the proposed method was the best for a missing ratio of 10\%, and SPCTV was the best for missing ratios of 50\%.
Both the proposed method and SPCTV were competitive for a missing ratio of 30\%.

\subsection{3D-volumetric image recovery}
In this experiments, we evaluated the proposed method using completion and denoising for 3D-volumetric magnetic resonance (MR) images.
Two MR images referred to as `citrus' and `tomato' have a size of (256 $\times$ 256 $\times$ 24).
As each voxel value is obtained using integration of a (1.5/8 mm $\times$ 1.5/8 mm $\times$ 1.5 mm)-volume, the resolutions of the (x,y,z)-axes are anisotropic.
Both MR images were corrupted by adding Gaussian noise $\sigma=1000$, and random voxels missing with various ratio $\rho \in \{ 10\%, 30\%, 50\%, 70\%, 90\% \}$.
For applying the proposed method, we consider $\bm w = [8/17, 8/17, 1/17]^T$ and $\bm \lambda = [1/3, 1/3, 1/3]^T$, and ($\alpha$,$\beta$,$\delta$) are tuned manually.
Table~\ref{tab:mri} shows the signal-to-distortion ratio (SDR), mean of SSIMs for slices, and computational times obtained using the proposed and state-of-the-art methods for MR image completion and denoising.
In computational times, LNRTC was the fastest, and the proposed method was the second.
In accuracy, SDR and SSIM of the proposed method outperformed the other methods for relatively low missing ratio: $\rho \in \{10\%, 30\%, 50\%\}$.
For high missing ratio: $\rho \in \{70\%, 90\%\}$, the SPCQV outperformed the other methods in SDR, and the proposed and GTV methods were very competitive outperforming the other methods
in SSIM.

\section{Conclusions}\label{sec:conclusion}
In this paper, we proposed a new model and algorithm for simultaneous tensor completion and denoising based on nuclear norms and TV minimization.
The proposed model can be characterized as a generalization of several prior works.
Our extensive experiments demonstrated the computational efficiency and the robustness to noise and missing of the proposed method.
Furthermore, this study revealed some creative usage of convex and non-convex methods such that our proposed method is good choice for highly noise cases, and non-convex methods (i.e., SPCQV/SPCTV) are good choices for high missing cases.

\section*{Acknowledgement}
This work was supported by Japan Society for the Promotion of Science KAKENHI Grant Number 15K16067.

\end{document}